\documentclass[10pt,twocolumn,letterpaper]{article}

\usepackage[pagenumbers]{cvpr} 
\usepackage{multirow}
\usepackage{bbding}
\usepackage{xcolor} 
\usepackage{graphicx}
\usepackage{amsmath}
\usepackage{amssymb}
\usepackage{booktabs}

\definecolor{cvprblue}{rgb}{0.21,0.49,0.74}
\usepackage[pagebackref,breaklinks,colorlinks,allcolors=cvprblue]{hyperref}

\def\ournet{ExpRDiff\xspace}

\title{ExpRDiff: Short-exposure Guided Diffusion Model \\for Realistic Local Motion Deblurring}

\author{Zhongbao Yang \quad Jiangxin Dong  \quad Jinhui Tang \quad Jinshan Pan \\
School of Computer Science and Engineering, Nanjing University of Science and Technology\\
{\tt \url{https://github.com/yzb1997/ExpRDiff}}
}

\begin{document}

\twocolumn[{
\maketitle
\vspace{-10mm}
\begin{center}
    \includegraphics[width=1\linewidth, height=0.29\textheight]{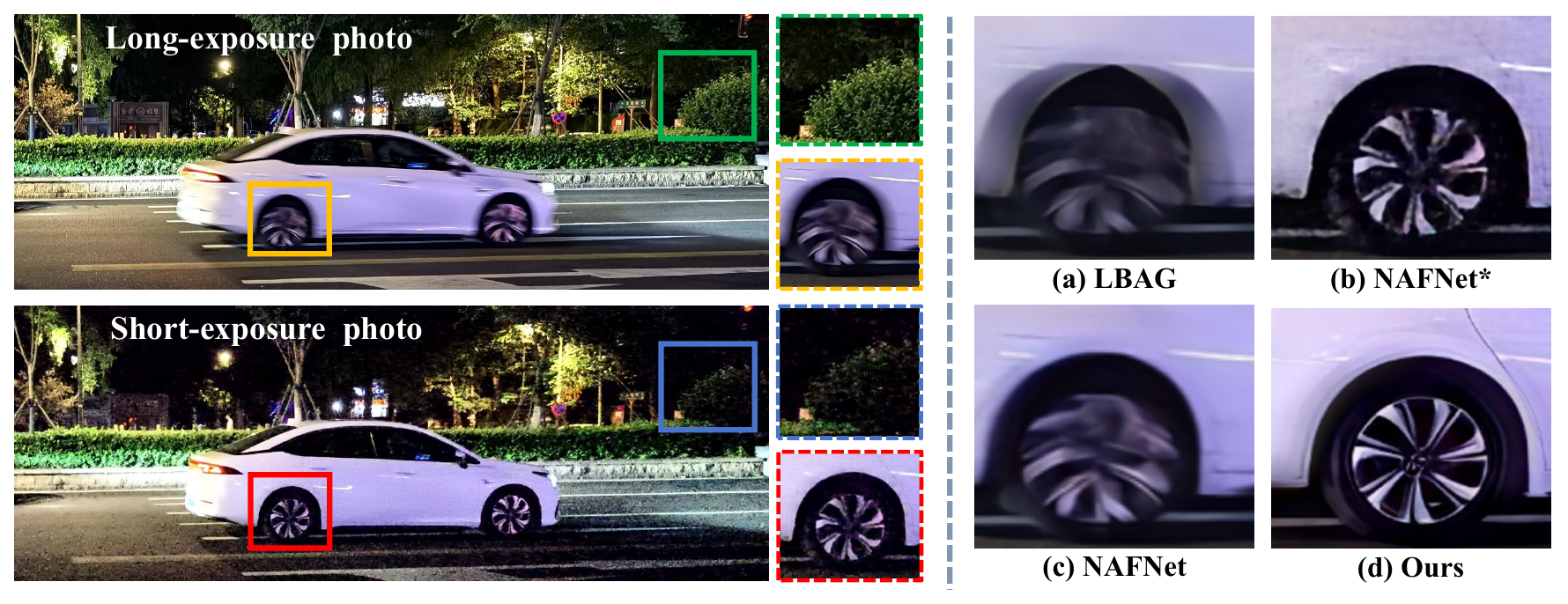} \\
    \captionof{figure}{Deblurring images with real-world local motion blur. The left side shows examples of a long-exposure photo (top) and a short-exposure photo (bottom) of the same scene captured by smart phones with vivo X100 pro. Both existing local blur removal methods, e.g., LBAG~\cite{ReLoBlur}, and state-of-the-art image deblurring methods, e.g., NAFNet~\cite{NAFNet}, do remove local blur well as shown in (a) and (c). In addition, simply using the short exposure image as reference does not solve this problem well, as shown in (b). In contrast to existing methods, we develop an effective \ournet to explore the sharp information from the short\-exposure images and blur information from the blurred images for blur removal and use them to guide the diffusion models for better realistic image restoration.}
    \label{fig:potential}
    \vspace{5mm}
\end{center}
}]

\begin{abstract}
Removing blur caused by moving objects is challenging, as the moving objects are usually significantly blurry while the static background remains clear.
Existing methods that rely on local blur detection often suffer from inaccuracies and cannot generate satisfactory results when focusing solely on blurred regions.
To overcome these problems, we first design a context-based local blur detection module that incorporates additional contextual information to improve the identification of blurry regions.
Considering that modern smartphones are equipped with cameras capable of providing short-exposure images, we develop a blur-aware guided image restoration method that utilizes sharp structural details from short-exposure images, facilitating accurate reconstruction of heavily blurred regions.
Furthermore, to restore images realistically and visually-pleasant, we develop a short-exposure guided diffusion model that explores useful features from short-exposure images and blurred regions to better constrain the diffusion process.
Finally, we formulate the above components into a simple yet effective network, named \ournet.
Experimental results show that \ournet performs favorably against state-of-the-art methods.
\end{abstract}
\vspace{-3mm}

\section{Introduction}
\label{sec:intro}
Existing smart phones equipped with anti-shake functions significantly improve the quality of captured images. However, capturing clear moving objects in a dynamic low-light environment is still a difficult problem as a longer exposure time is often required and the speed of the moving objects is fast. The long-exposure photo with significant local blur effects; e.g., the objects are blurry while the static background is usually clear. Thus, how to deblur images with the local blur effect receives significant attention from both academic and industry communities.
Most existing methods are designed mainly for images with a uniform or non-uniform global blur effect~\cite{MIMO, Deeprft, MPRNet, NAFNet, Uformer, Fftformer, Stripformer, Restormer}. When applying these methods to images with local blur, the clear background will be over smoothed, while moving objects are not restored well. Thus, it is of great need to develop an effective method to handle the clear background and blurry objects simultaneously for better image restoration.
To this end, several approaches~\cite{ReLoBlur, LMD-ViT} adopt object segmentation or blur detection approaches to differentiate between blurry and clear regions, subsequently handling blur regions and clear regions simultaneously. However, local blurs resulting from object motion often present significant challenges (e.g., the tire in the blue boxes of the $2^{st}$ patch in Figure~\ref{fig:potential}), and adverse imaging conditions, such as low-light environments, further increase the difficulty of object segmentation or blur detection. As a result, reliance solely on object segmentation or blur detection does not effectively improve deblurring performance.
We note that existing handheld devices can capture multiple photos of a moving object simultaneously through burst mode or multi-camera systems. Photos captured using these settings typically contain sharp details due to the short-exposure times (e.g., the tire in the yellow boxes of the $2^{st}$ patch in Figure~\ref{fig:potential}). Thus, it is of great interest to explore the information of these images to facilitate local blur removal.
To solve the problems mentioned above, we develop an \ournet to explore information from localized blur regions and short-exposure images as guidance to facilitate local blur removal.
First, given that not all the regions of captured images are blurry, we develop a context-based local blur detection module to explore contextual information for better identification of the blurry regions.
Then, we develop a simple yet effective blurry-aware guided image restoration that explores the useful information of short-exposure images and the detected blurry regions as guidance to discriminately handle the blurry regions and clear regions simultaneously for better image restoration.
However, as the moving speeds of the objects are usually fast, which leads to significant blur effects, restoring realistic clear images with rich details from heavily blurred images is not a trivial task.
Motivated by the success of diffusion-based methods~\cite{SUPIR, PASD} in high-quality image generation, we take advantage of diffusion priors to help with realistic clear image restoration.

To ensure that the diffusion model pays more attention to restoring reliable structural details, we develop a simple yet effective feature fusion module, named ExpBFusion, to adaptively extract useful features from the short-exposure images and the detected blurry regions. These features obtained through ExpBFusion can then guide the diffusion model for better image restoration.
By formulating the aforementioned modules into a unified framework, the proposed \ournet achieves favorable performance against state-of-the-art methods on publicly available local blur datasets as well as real-world scenarios.
The main contributions are summarized as follows:
\begin{itemize}
    \item We develop a context-based local blur detection module to explore additional contextual information to improve the identification of blurry regions.
    \item We develop a blurry-aware guided image restoration to remove local blur by handling the blurry regions and clear regions discriminately in the restoration process.
    \item We develop a short-exposure guided diffusion model to explore useful features from the short-exposure images and the detected blurry regions as guidance for realistic image restoration.
\end{itemize}
%

%
\begin{figure*}[htbp!]
    \centering
 \includegraphics[width=1\textwidth]{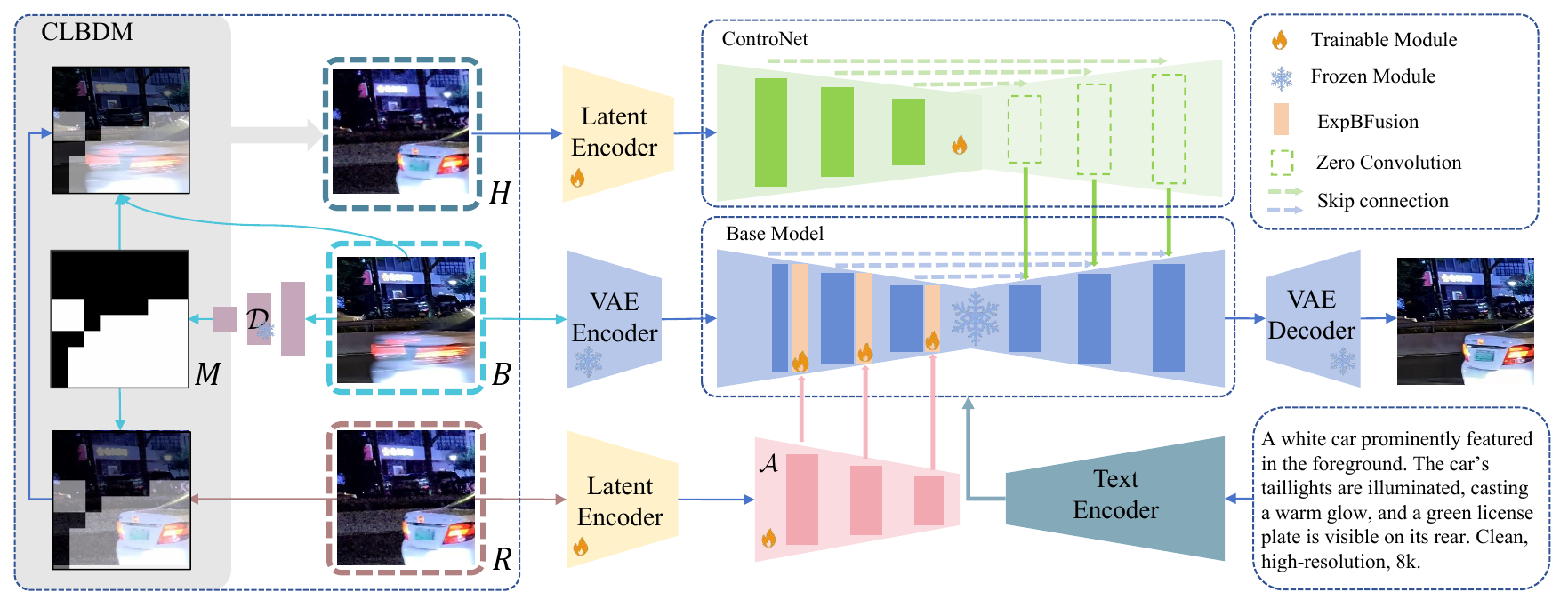}
 \vspace{-4mm}
 \caption{
 An overview of the proposed method ExpRDiff. $\mathcal{A}$ is the network to extract features for $R$. $\mathcal{D}$ denotes Blurry-aware guided image restoration, which aims to extract short-exposure image features, CLBDM denotes the context-based local blur detection module. $B$, $R$, $M$, and $H$ are blurry images, short-exposure images, confidence maps, and the result of CLBDM respectively.}
 \label{fig: Network FlowChart}
\vspace{-1mm}
\end{figure*}
%
%
\section{Related Work}
\label{sec:related}
{\flushleft \textbf{Hand-crafted prior-based methods.}} Since image deblurring is ill-posed, conventional methods usually use reference images with clearer structural information as effective priors to jointly solve image deblurring.
In~\cite{RDMD}, Chen et al. propose joint image deblurring with two images with different levels of blurring.
Zhuo et al.~\cite{RFD} and Yuan et al.~\cite{IDBNIP} to perform image deblurring with the help of flash images and noisy images as reference images, in which the structure is clearer, respectively.
Although traditional joint image deblurring methods can achieve better results than traditional single image deblurring methods, traditional joint image deblurring methods assume that the blurring are global of these images can be well described by a single blur kernel result from camera shake and the scene is static.
These methods provide poor performance of deblurring in images with non-uniform blurring and blurred images caused by object motion.
{\flushleft \textbf{Deep learning-based methods.}}
To solve more complex image deblurring situations avoiding the limitations assumed in manually designing image priors, Deep learning-based image deblurring methods are explored to solve image deblurring.
Lost of single image deblurring methods ~\cite{SRN, DeblurGAN, DeblurGANv2, DMPHN, SAPHN, MIMO, Deeprft, Uformer, Stripformer} achieve favorable performance.
Zamir et al.~\cite{MPRNet} propose a multi-stage progressive image deblurring method for the restoration image. The non-parameter sharing network is utilized in different stages to learn the deblurring function of the corresponding scales.
While this design achieves competitive performance, the complexity of the model thus becomes larger.
To reduce the parameters of the model, Chen et al.~\cite{NAFNet} propose a computationally efficient simple baseline, which considers the nonlinear activation functions to be non-essential and replaces them.
But the amount of convolutional operations makes those CNNs-based methods only have the ability to capture localized information.
With the development of transformer~\cite{Transformer} in high-level vision tasks, whose ability to capture global contexts caught the attention to researchers.
~\cite{Restormer} calculates channel attention and ~\cite{Fftformer} computes the dot-product attention on the spatial frequency domain separately, which reduces the complexity of computing the attention weights in the transformer.
~\cite{LMD-ViT} propose a window-based local motion deblurring vision transformer network, which employs the confidence predictor that constrains the output at the pixel level using the mask of gt to predict blur areas.
The confidence predictor network is trained in such kind of strategy that it leads to mask generation inaccurately.
%
%
{\flushleft \textbf{Diffusion-based image restoration.}}
With the surprising performance of StableDiffusion-XL (SDXL)~\cite{SDXL} in image image-generating field, it has been developed for image restoration.
To solve the problem of uncontrollable generation of images in the image generation task, ~\cite{T2I_adapter, controlnet} employ another branch to extract the features of the condition to control the final generated output.
Inspired by that, \cite{PASD, SUPIR, DiffBIR} utilize the degradation network to generate clear images or clear latent from the input degraded images, which is used to put into the denoising model and the ControlNet~\cite{controlnet} to control the diffusion process and generate high-quality images.
Although the images recovered by those methods achieve pleasant visualization results, the training of these methods is based on a large collection of high-quality images and utilizes synthetic degraded images, which do not allow for the restoration of local blur caused by object motion.
In order to better utilize the prior provided by SDXL to recover local blur in the presence of a reference image, we propose the diffusion-based low-light local motion deblurring network.
\section{Proposed Method}
\label{sec:method}
The proposed \ournet aims to explore useful features from short-exposure images and blurry regions as guidance for local blur removal and realistic image restoration. It contains a context-based local blur detection module that is used to identify blurry regions by exploring contextual information, a blurry-aware guided image restoration that uses the short-exposure images and blurry regions to remove local blur, and a short-exposure guided diffusion model that is constrained by the features of short-exposure images and blurry regions for realistic image restoration. Figure~\ref{fig: Network FlowChart} shows the overview of the proposed \ournet. In the following, we explain each component in detail.
\vspace{-2mm}
\begin{figure*}[htbp!]
    \centering
 \includegraphics[width=1\textwidth]{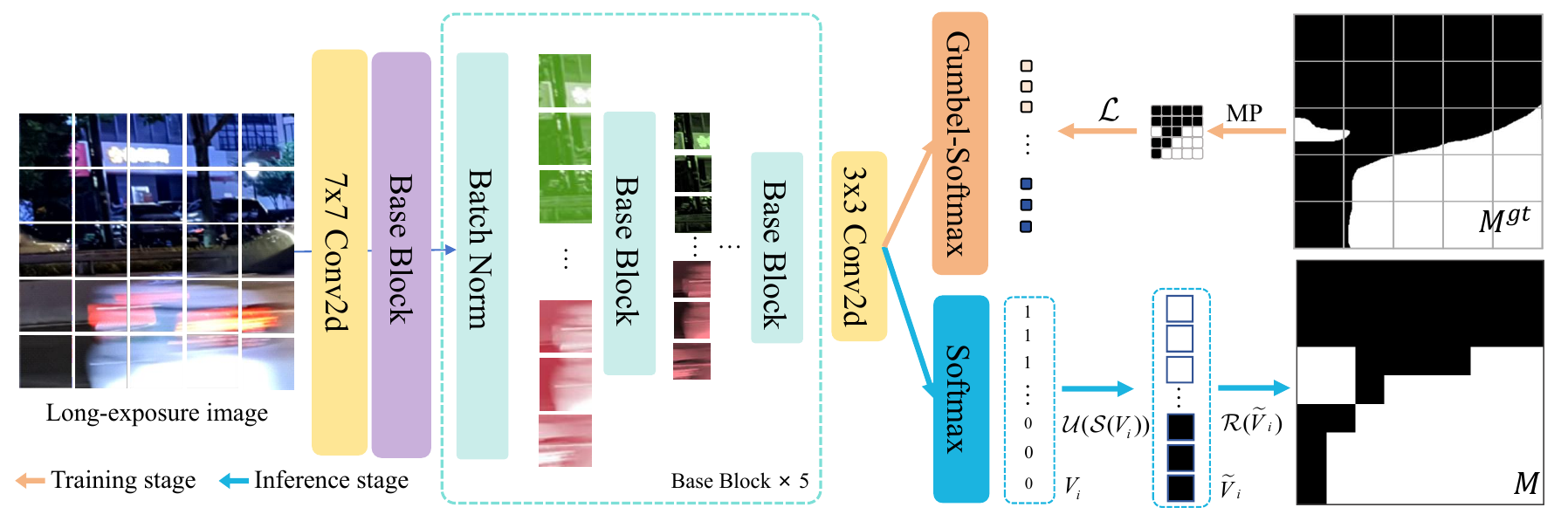}
 \vspace{-4mm}
 \caption{
     An overview of the proposed context-based local blur detection module to detect the blur regions of the long-exposure blur image. MP denotes the Max-pooling operation.}
 \label{fig: Network FlowChart3}
 \vspace{-3mm}
\end{figure*}
\subsection{Context-based local blur detection module}
\vspace{-2mm}
Local blur detection is an essential step for handling images with local blur. Existing methods~\cite{ReLoBlur, LMD-ViT} usually treat this problem as a binary classification and employ a binary network to determine whether each pixel position is blurry or not.
However, this approach, which relies solely on individual pixel characteristics, often fails to capture contextual information, leading to inaccurate blur detection and subsequent degradation in overall result quality.
To address this problem, we develop a simple yet effective context-based local blur detection method that incorporates additional contextual information to enhance the robustness of local blur detection.

Specifically, we first divide the input blurry images into $N$ non-overlapping patches $\{P_{i}\}^{N}_{i=1}$ with the unfolding operation from the input blurry image $B$, where $P_{i} \in \mathbb{R}^{C\times60\times60}$.
Then, we develop a simple yet effective network $\mathcal{D}(\cdot)$ to estimate a set of confidence values $\{V_{i}\}^{N}_{i=1}$, $V_{i} \in \mathbb{R}^{C\times1\times1}$, from the input blurry image to discriminate whether the divided patches $\{P_{i}\}^{N}_{i=1}$ are blurry or not:
\begin{equation}
    V_{i} = \mathcal{S}\left(\frac{\mathcal{D}(P_{i}) + G_{i}}{\tau}\right),
\end{equation}
where $G_{i} = -\log\left(-\log(U_{i})\right)$, $\quad U_{i} \sim \text{Uniform}(0, 1)$, $\mathcal{S}(\cdot)$ denotes the Softmax operation, $\tau$ is the temperature parameter that controls the smoothness of the distribution, $G_{i}$ represents the Gumbel noise added for differentiable sampling~\cite{gumbelsoftmax}, and $U_{i}$ is a random variable sampled from a uniform distribution $\text{Uniform}(0, 1)$.
Figure~\ref{fig: Network FlowChart4} illustrates the detailed architecture of the module.

We train the network $\mathcal{D}$ by minimizing the following Cross-Entropy loss function:
\begin{equation}
    \mathcal{L} = -\frac{1}{N} \sum_{i=1}^{N} \left[ V_{i} \log(M_{i}^{gt}) + (1 - V_{i})\log(1 - M_{i}^{gt}) \right],
\end{equation}
where $M_{i}^{gt}$ denotes the ground truth confidence map that indicates the blur degree of the patch $P_i$. It is obtained by cropping the corresponding region from the ground truth blur confidence map $M^{gt}$ based on the spatial position of patch $P_{i}$.

As the confidence map is estimated in image patches, we employ the nearest-neighbor interpolation to obtain the pixel-level confidence map $M$ to discriminate the blurry regions and clear regions from input image:
\begin{equation}
\begin{split}
    &\Tilde{V}_i = \mathcal{U}(\mathcal{S}(V_i)), \quad V_{i} \in \mathbb{R}^{C\times60\times60}, \\
    &M = \mathcal{R}(\Tilde{V}_i),
    \end{split}
\end{equation}
where $\mathcal{U}(\cdot)$ denotes the nearest-neighbor interpolation operation, $\mathcal{R}(\cdot)$ denotes the folding operation.
In addition to better detection of blurred regions, the confidence map generated by the proposed context-based local blur detection module further helps short-exposure images guide the recovery of realistic effects in blurred regions.
We will show its effectiveness in Section~\ref{ablation}.
\subsection{Blurry-aware guided image restoration}
\label{sec:MGIF}
\begin{figure}[]
\footnotesize
\centering
    \begin{tabular}{cccc}
    \hspace{-2mm} \includegraphics[width=0.11\textwidth]{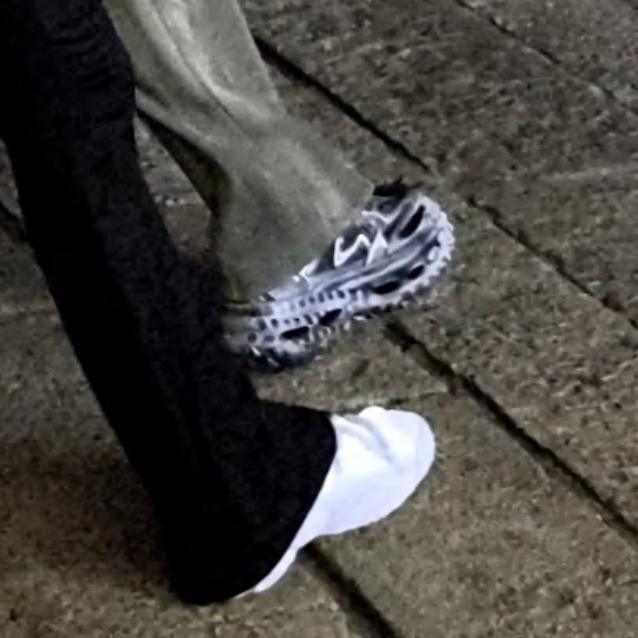}
    & \hspace{-3mm} \includegraphics[width=0.11\textwidth]{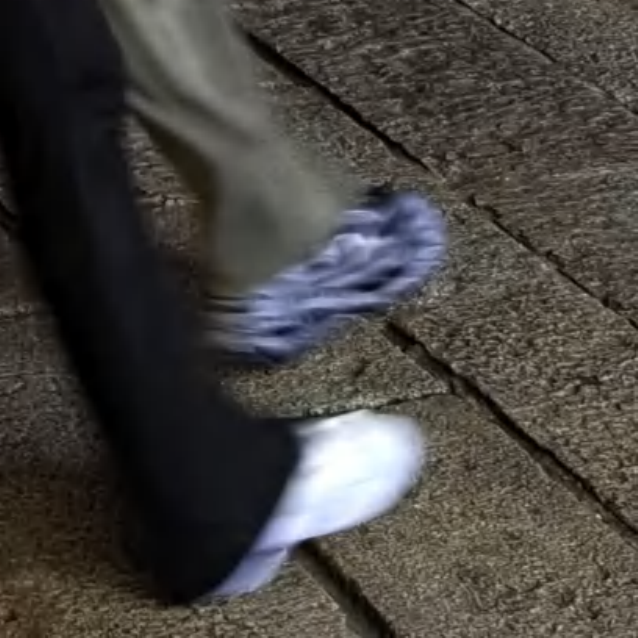}
    & \hspace{-3mm} \includegraphics[width=0.11\textwidth]{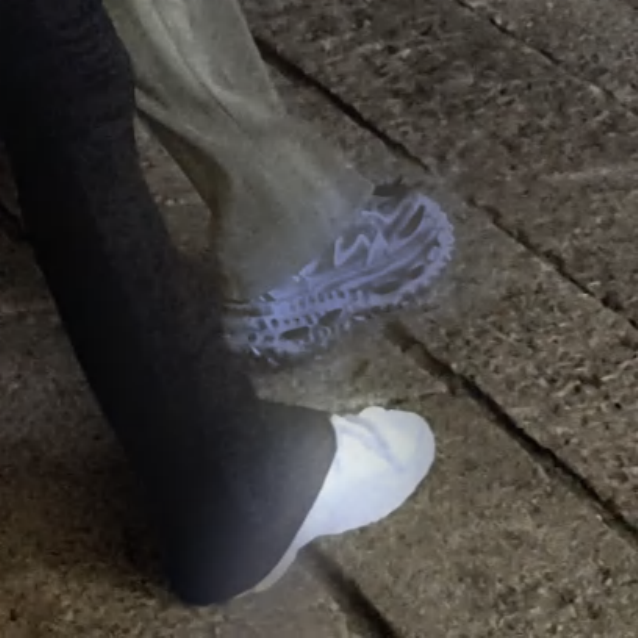}
    & \hspace{-3mm} \includegraphics[width=0.11\textwidth]{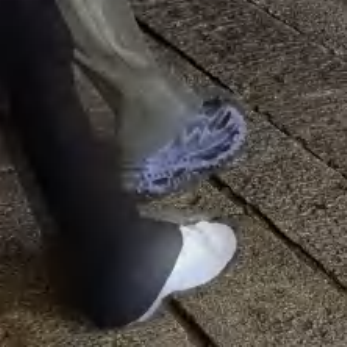} \\
    \hspace{-2mm}(a)  &\hspace{-2mm} (b)  &\hspace{-5mm} (c)  &\hspace{-5mm} (d) \\
    \end{tabular}
    \vspace{-1mm}
    \caption{Illustration of guided image filtering~\cite{GIF}. (a) Short-exposure reference; (b) blurred input; (c) result of direct guided filtering, which reduces blur but smooths the static background; (d) result of masked guided filtering, preserving background clarity while achieving the same deblurring effect in blurred regions as (c).}

    \label{fig:gf}
    \vspace{-5mm}
\end{figure}

With the guidance of a short-exposure image, one can use the guided filtering method~\cite{GIF} to help eliminate blur. However, directly using~\cite{GIF} will smooth the clear background as shown in Figure~\ref{fig:gf}.
Meanwhile, the short-exposure images in clear regions have poor quality in low-light environments (e.g., the tree in the blue boxes of the $2^{st}$ patch in Figure~\ref{fig:potential}).
A natural solution to this problem is that we do not handle the clear regions of the input images.
To this end, we adopt a threshold based on the default value of the Gumbel-Softmax distribution to binarize $M$.
Then, we develop a blurry-aware guided restoration approach that formulates the guided filtering method as:
\begin{equation}
     H(\textrm{x}) = M(\textrm{x})(a_kR(\textrm{x})+b_k) + (1-M(\textrm{x}))B(\textrm{x}), \textrm{x} \in w_k,
     \label{modified GF}
\end{equation}
where $a_k$ and $b_k$ are linear coefficients in an image patch $w_k$; $B(\textrm{x})$ and $R(\textrm{x})$ denote the input blurry image and short-exposure image at pixel $\textrm{x}$.

The linear coefficients $a_k$ and $b_k$ can be obtained by solving the following optimization problem:
\begin{small}
\begin{equation}
     \min_{a_k, b_k}\sum_{i\in w_{k}}(M(\textrm{x})(a_kR(\textrm{x})+b_k) + (1-M(\textrm{x}))B(\textrm{x})-B(\textrm{x}))^{2}+\epsilon a^{2}_{k}),
     \label{objective function}
\end{equation}
\end{small}
where $\epsilon$ is a weight parameter. Similar to~\cite{GIF}, the solution of $a_{k}$ and $b_{k}$ can be obtained:
\begin{small}
\begin{equation}
    \begin{split}
        &a_k = \frac{\sum_{i \in w_k} M(\textrm{x}_i)^2 R(\textrm{x}_i) B(\textrm{x}_i) - \sum_{i \in w_k} M(\textrm{x}_i)^2 R(\textrm{x}_i) b_k}{\sum_{i \in w_k} M(\textrm{x}_i)^2 R(\textrm{x}_i)^2 + \epsilon},\\
        &b_k = \frac{\sum_{i \in w_k} M(\textrm{x}_i)^2 B(\textrm{x}_i) - \sum_{i \in w_k} M(\textrm{x}_i)^2 a_k R(\textrm{x}_i)}{\sum_{i \in w_k} M(\textrm{x}_i)^2}.
    \end{split}
    \label{eq: 6}
\end{equation}
\end{small}
The details about the derivation of Eqs.~(\ref{eq: 6}) are included in the supplemental material.
\subsection{Short-exposure guided diffusion model}
\label{SGDM}
Although the proposed blurry-aware guided image restoration can remove local blur well, it is less effective to restore realistic images with details due to the significant blur caused by moving objects.
We note that existing diffusion-based methods with text prompts can recover details from the degraded image.
However, text information is a high-level feature and cannot effectively characterize blurry regions and low-level structural features of images. Thus, only using the text prompt does not guide the diffusion models for realistic image restoration when local blur exists.
To address this problem, we propose a short-exposure guided diffusion model that explores the features of both short-exposure images and blurry regions for better image restoration.
\begin{figure}[tb]
    \centering
 \includegraphics[width=0.45\textwidth]{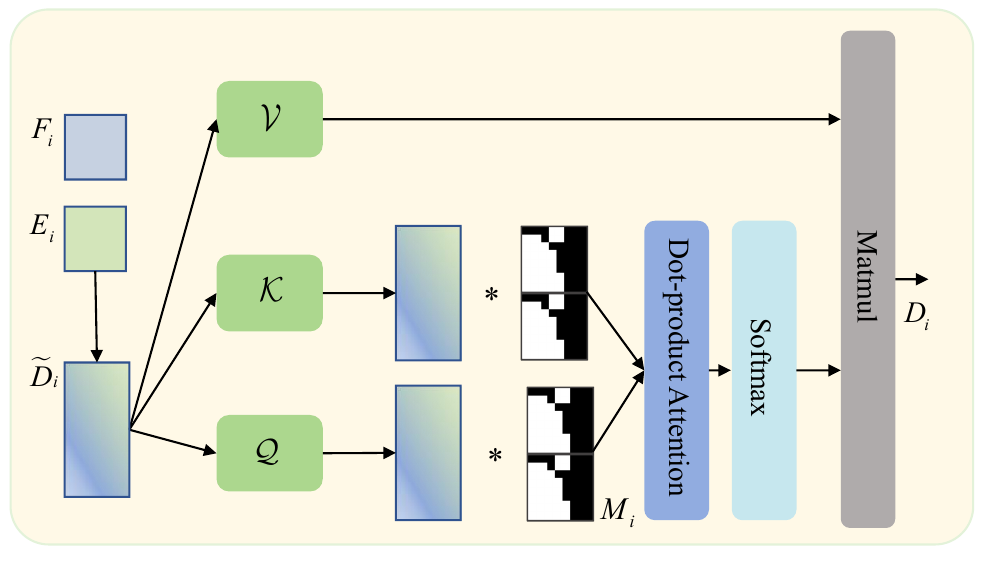}
 \caption{
 Details of the proposed ExpBFusion module.}
 \label{fig: Network FlowChart4}
\end{figure}

Specifically, assuming that the multi-scale features $\{E_i\}_{i=1}^s$ are from the encoder module of the U-Net in the diffusion model, where $s$ denotes the number of scales, we first extract the multi-scale features $\{F_i\}_{i=1}^s$ from the short-exposure image by a feature extraction network\footnote{The details about the feature extraction network are included in the supplemental material.}, where the spatial resolution of $F_i$ is the same as $E_i$.
Then, with the multi-scale features $\{F_{i}\}_{i=1}^{s}$ and the estimated confidence map $M$, we develop ExpBFusion, a simple yet effective feature fusion module, to guide $\{E_{i}\}_{i=1}^{s}$ in the diffusion process:
\begin{equation}
\begin{split}
    D_i = \mathcal{S}\left(\frac{\mathcal{Q}(\Tilde{D}_i)\mathcal{K}(\Tilde{D}_i)^\top}{\gamma}\right)\mathcal{V}(\hat{D}_i),
    \label{eq:7}
    \end{split}
\end{equation}
where $\Tilde{D}_i = \mathcal{C}(F_i\otimes M_i, E_i\otimes M_i)$ and $\hat{D}_i = \mathcal{C}(F_i, E_i)$; $\mathcal{C}(\cdot)$ denotes a concatenation operation along the spatial dimension; $M_i$ denotes the nearest downsmpling result of $M$, which has the same spatial resolution as $F_i$; $\otimes$ denotes the element-wise product operation; $\mathcal{Q}(\cdot)$, $\mathcal{K}(\cdot)$, and $\mathcal{V}(\cdot)$ denote the operations for extracting the query, key, and value matrices, respectively; $\gamma$ is a learnable weight parameter.
Finally, we use the fused feature $D_i$ instead of $E_i$ in the diffusion process.
The detailed module is shown in Figure~\ref{fig: Network FlowChart4} and we employ the base model of StableDiffusion-XL (SDXL)~\cite{SDXL} as our diffusion model as it generates high-quality and realistic images.
We refer to the module by Eq.~(\ref{eq:7}) as the proposed short-exposure guided diffusion model and show its effectiveness and efficiency in Section~\ref{ablation}.

\begin{figure*}[!t]
\footnotesize
\centering
    \begin{tabular}{c c c c c c c}
            \multicolumn{3}{c}{\multirow{5}*[45.6pt]{
            \hspace{-2.5mm} \includegraphics[width=0.325\linewidth,height=0.235\linewidth]{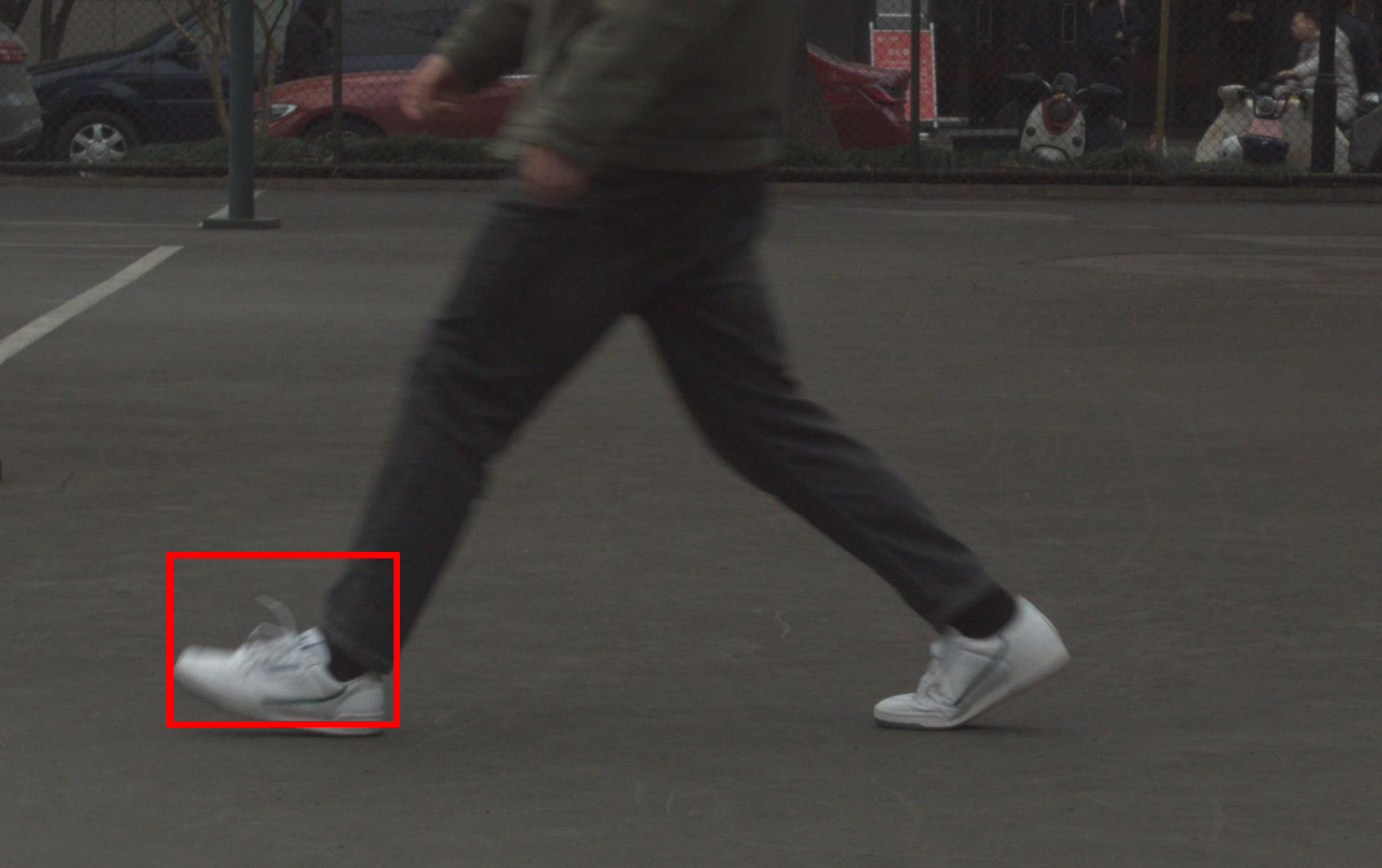}}}
            & \hspace{-4.0mm} \includegraphics[width=0.16\linewidth,height=0.105\linewidth]{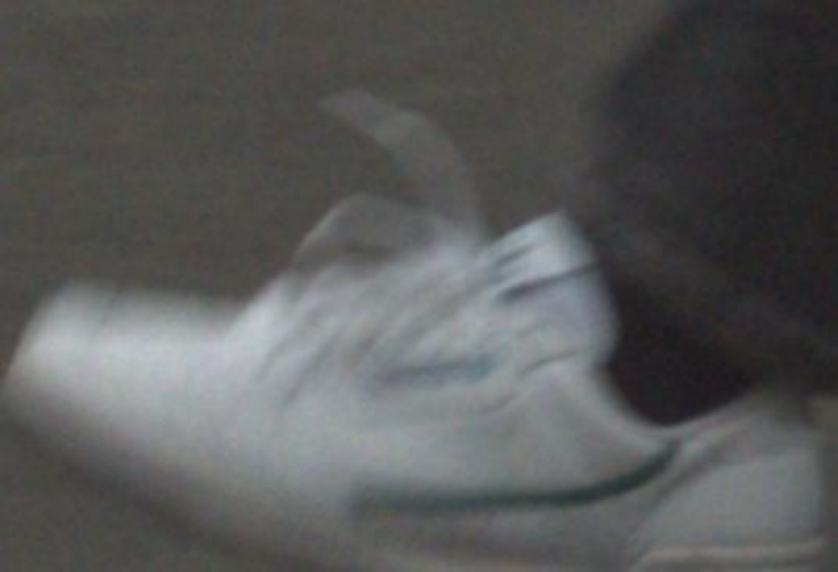}
            & \hspace{-4.0mm} \includegraphics[width=0.16\linewidth,height=0.105\linewidth]{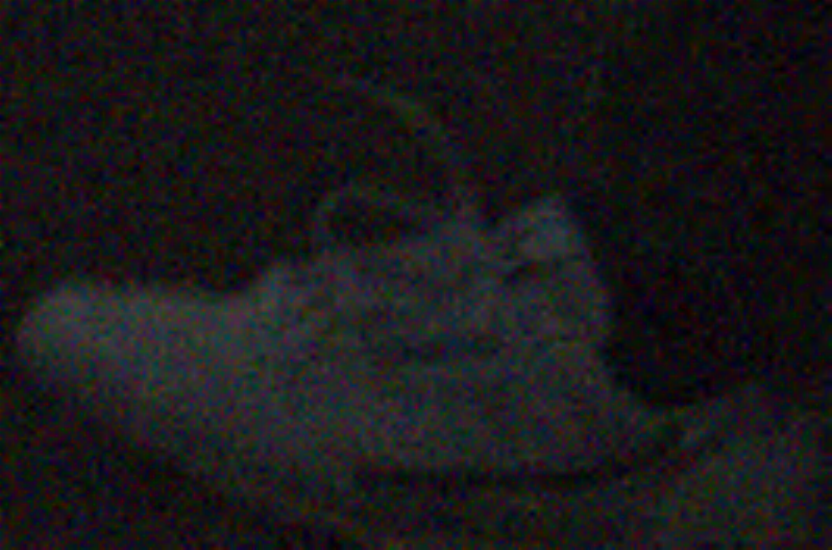}
            & \hspace{-4.0mm} \includegraphics[width=0.16\linewidth,height=0.105\linewidth]{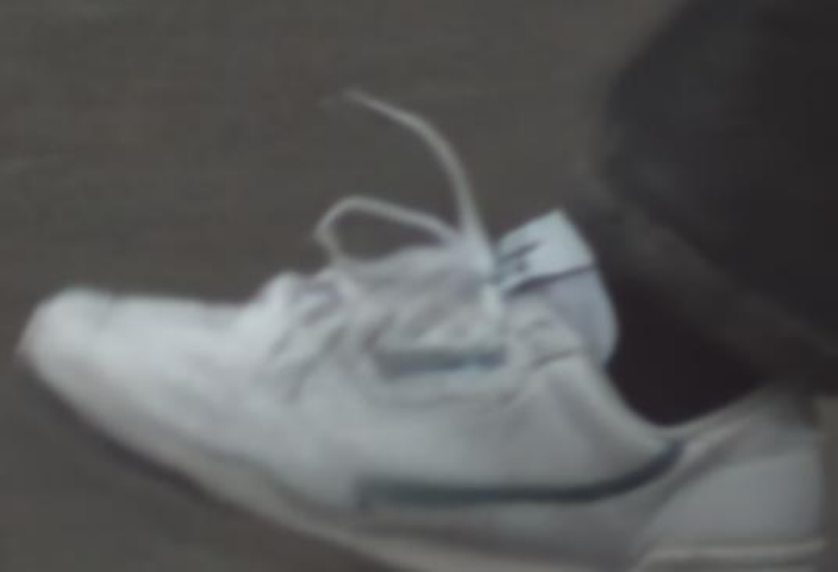}
            & \hspace{-4.0mm} \includegraphics[width=0.16\linewidth,height=0.105\linewidth]{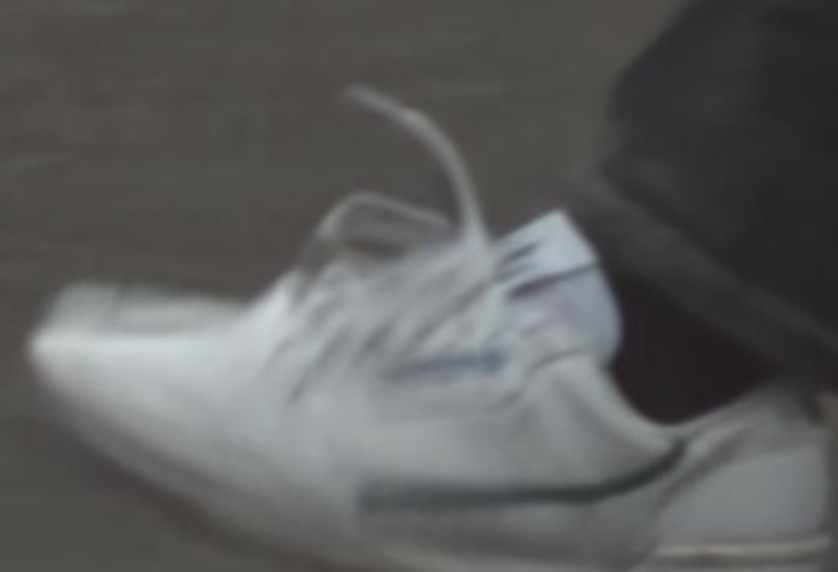}

              \\
    		\multicolumn{3}{c}{~}
            & \hspace{-4.0mm} (a) Blurred patch
            & \hspace{-4.0mm} (b) Short-exposure patch
            & \hspace{-4.0mm} (c) LMD-ViT~\cite{LMD-ViT}
            & \hspace{-4.0mm} (d) NAFNet~\cite{NAFNet} \\	

    	\multicolumn{3}{c}{~}
            & \hspace{-4.0mm} \includegraphics[width=0.16\linewidth,height=0.105\linewidth]{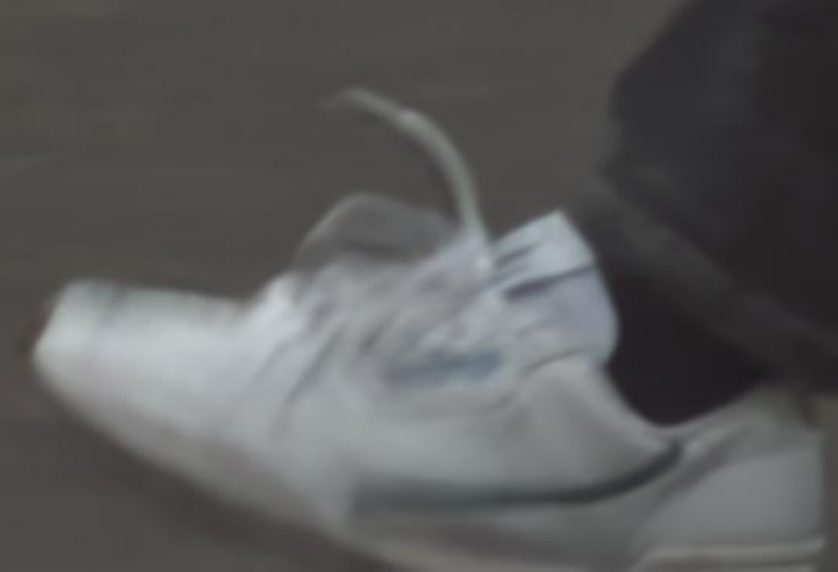}
            & \hspace{-4.0mm} \includegraphics[width=0.16\linewidth,height=0.105\linewidth]{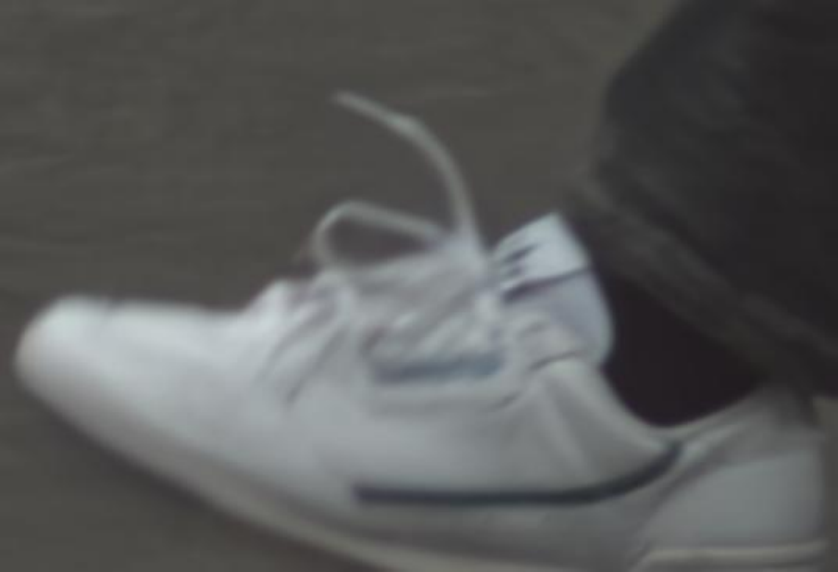}
            & \hspace{-4.0mm} \includegraphics[width=0.16\linewidth,height=0.105\linewidth]{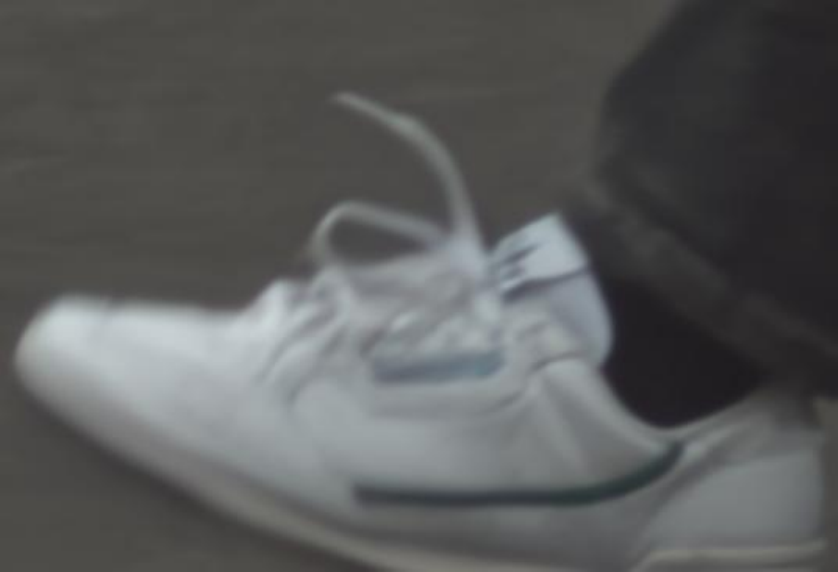}
            & \hspace{-4.0mm} \includegraphics[width=0.16\linewidth,height=0.105\linewidth]{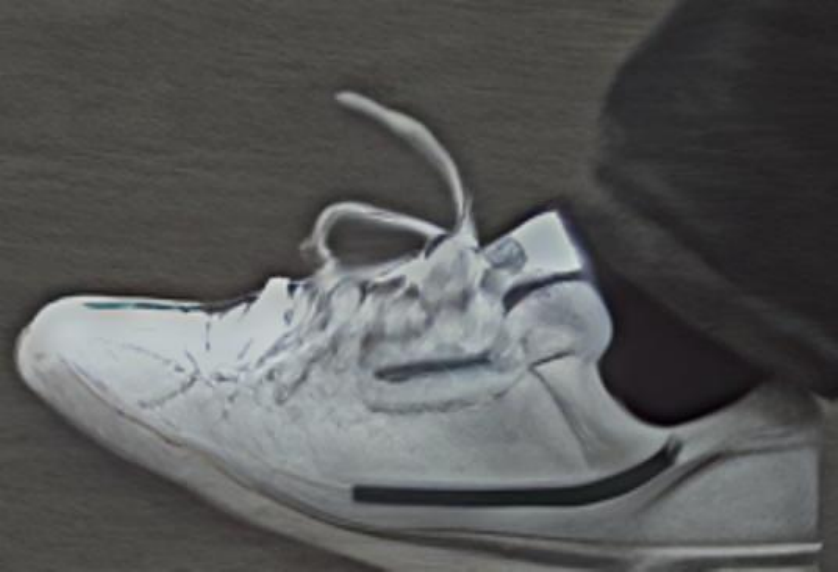}
            \\

    	\multicolumn{3}{c}{\hspace{-4.0mm} \textbf{Blurry input}}
            & \hspace{-4.0mm} (e) Restormer~\cite{Restormer}
            & \hspace{-4.0mm} (f) NAFNet*~\cite{NAFNet}
            & \hspace{-4.0mm} (g) Restormer*~\cite{Restormer}
            & \hspace{-4.0mm} (h) Ours
            \\
            \multicolumn{3}{c}{\multirow{5}*[45.6pt]{
            \hspace{-2.5mm} \includegraphics[width=0.325\linewidth,height=0.235\linewidth]{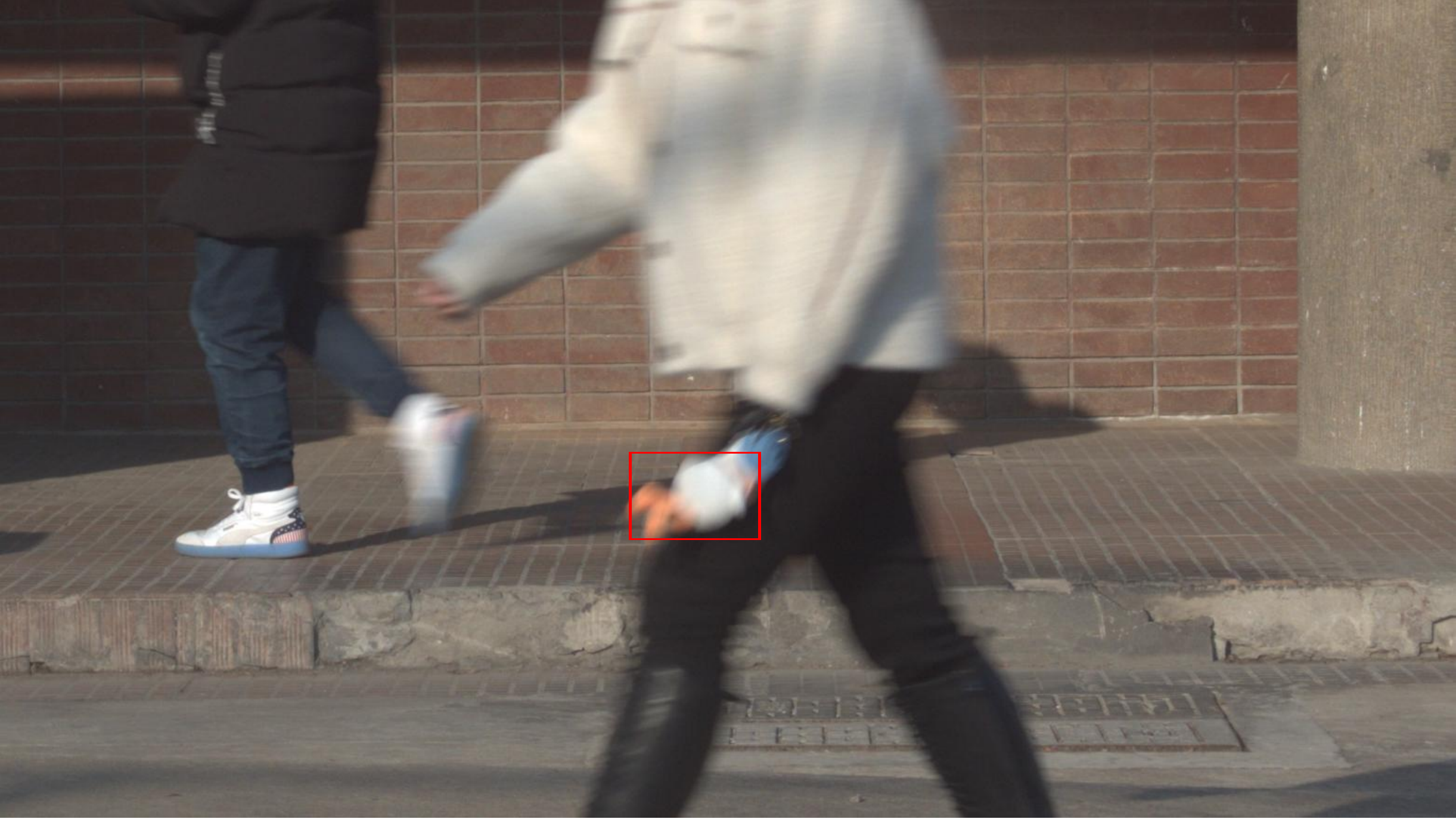}}}
            & \hspace{-4.0mm} \includegraphics[width=0.16\linewidth,height=0.105\linewidth]{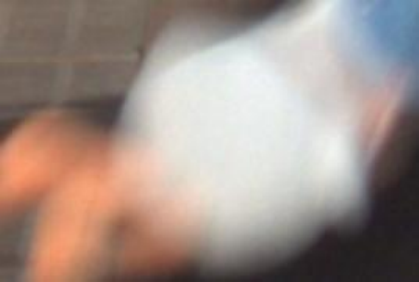}
            & \hspace{-4.0mm} \includegraphics[width=0.16\linewidth,height=0.105\linewidth]{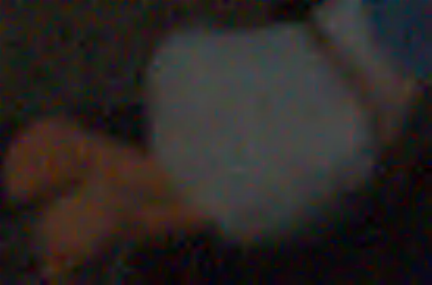}
            & \hspace{-4.0mm} \includegraphics[width=0.16\linewidth,height=0.105\linewidth]{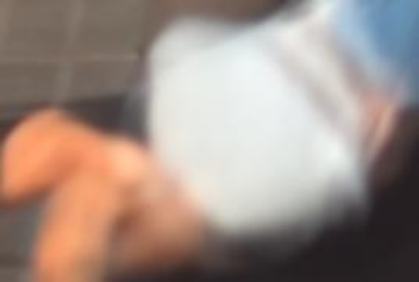}
            & \hspace{-4.0mm} \includegraphics[width=0.16\linewidth,height=0.105\linewidth]{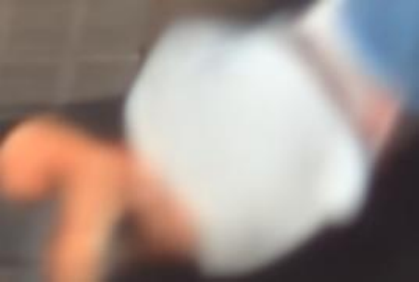}

              \\
    		\multicolumn{3}{c}{~}
            & \hspace{-4.0mm} (a) Blurred patch
            & \hspace{-4.0mm} (b) Short-exposure patch
            & \hspace{-4.0mm} (c) LMD-ViT~\cite{LMD-ViT}
            & \hspace{-4.0mm} (d) NAFNet~\cite{NAFNet} \\	

    	\multicolumn{3}{c}{~}
            & \hspace{-4.0mm} \includegraphics[width=0.16\linewidth,height=0.105\linewidth]{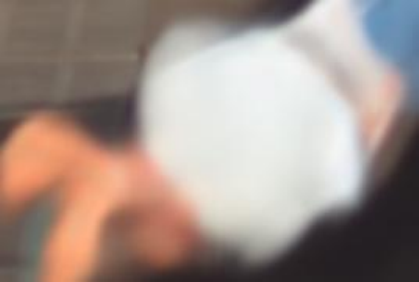}
            & \hspace{-4.0mm} \includegraphics[width=0.16\linewidth,height=0.105\linewidth]{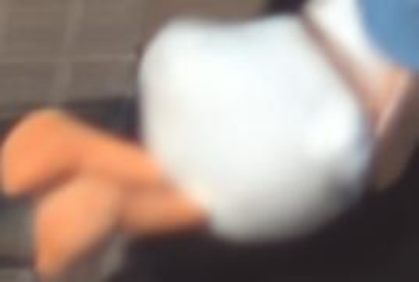}
            & \hspace{-4.0mm} \includegraphics[width=0.16\linewidth,height=0.105\linewidth]{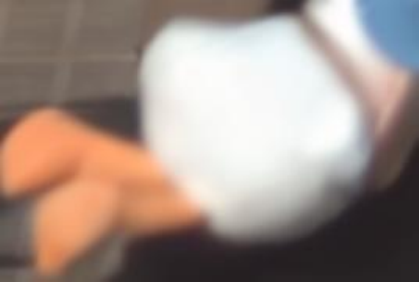}
            & \hspace{-4.0mm} \includegraphics[width=0.16\linewidth,height=0.105\linewidth]{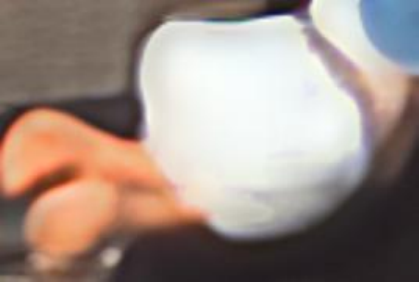}
            \\

    	\multicolumn{3}{c}{\hspace{-4.0mm} \textbf{Blurry input}}
            & \hspace{-4.0mm} (e) Restormer~\cite{Restormer}
            & \hspace{-4.0mm} (f) NAFNet*~\cite{NAFNet}
            & \hspace{-4.0mm} (g) Restormer*~\cite{Restormer}
            & \hspace{-4.0mm} (h) Ours
            \\

    \end{tabular}
\vspace{-1mm}
\caption{Deblurred results on the ReLoblur synthetic dataset~\cite{ReLoBlur}. The deblurred results in (c)-(g) still contain significant blur effects. The proposed method generates a better deblurred image with clearer structures in (h).}
\label{fig:visual1}
\vspace{-3mm}
\end{figure*}

\section{Experimental Results}
We evaluate the effectiveness of the proposed method against the state-of-the-art ones using public benchmark datasets.
%
\subsection{Experimental settings}
{\flushleft \textbf{Training datasets.}}
We use the ReLoBlur dataset~\cite{ReLoBlur} as our training dataset.
To generate short-exposure images, we first convert the sharp images into the hue saturation value (HSV) color space to reduce the brightness and add random noise to these images.
Then we crop the input images and the corresponding masks with the spatial size of $512 \times 512$ pixels to focus on local details.
We split the dataset into 19,351 pairs for training and 3,824 for evaluation.
To demonstrate the generalization ability of our proposed method on real-world data, we capture the real local motion blurred dataset (LMBD) that contains 10,380 images for training and 209 images for evaluation. The details are included in the supplemental material.
{\flushleft \textbf{Parameter settings.}}
We implement our method using the PyTorch framework and train it on a machine with 8 NVIDIA GeForce RTX 4090 GPUs.
The AdamW~\cite{adamw} optimizer with default parameters is used. The patch size is set to be $256 \times 256$ pixels, and the batch size is 32.
We employ the loss function~\cite{PASD} to regularize our network and train the proposed network with 160,000 iterations, The initial learning rate is set to $5 \times 10^{-5}$.
\begin{table}[!ht]\footnotesize
\caption{Quantitative evaluations of the proposed approach against state-of-the-art methods on the ReLoBlur synthetic dataset~\cite{ReLoBlur}. The ``*'' represents that we concat the $R$ and $B$ as the model input. The $\downarrow$'' and $\uparrow$'' represent that lower values indicate better results, and higher values indicate better results, respectively.}
    \begin{tabular}{lcccc}
    \hline
    Methods                     &NIQE $\downarrow$          &CLIPIQA+$\uparrow$           &MUSIQ-PAQ2PIQ $\uparrow$        \\ \hline
    MIMO-UNet~\cite{MIMO}       &6.1947                     &0.3525             &58.0449              \\
    NAFNet~\cite{NAFNet}        &5.9387                     &0.3631             &59.9376              \\
    Restormer~\cite{Restormer}  &6.1146                     &0.3596             &59.2634              \\
    MIMO-UNet*~\cite{MIMO}      &5.9514                     &0.3607             &60.9325              \\
    NAFNet*~\cite{NAFNet}       &5.8419                     &0.3644             &61.2459              \\
    Restormer*~\cite{Restormer} &5.8826                     &0.3669             &61.4244              \\
    LBAG~\cite{ReLoBlur}        &5.9243                     &0.3644             &60.1322              \\
    LMD-ViT~\cite{LMD-ViT}      &5.9341                     &0.3389             &59.8619              \\
    Ours                        &\textbf{5.2212}            &\textbf{0.3709}    &\textbf{65.8149}     \\ \hline
\end{tabular}
\label{compare1}
\vspace{-4mm}
\end{table}
{\flushleft \textbf{Evaluation metrics.}}
How to better evaluate the quality of restored images is important. Note that conventional distortion measures, e.g., PSNR and SSIM do not measure the visual quality of images well~\cite{tpdp}. We mainly use NIQE, CLIPIQA+, and MUSIQ-PAQ2PIQ as the main evaluation metrics as they are able to measure the visual quality of images well.
\subsection{Comparisons with the state of the art}
\begin{table}[!ht]\footnotesize
\caption{Quantitative evaluations on the LMBD dataset. All the comparison results are generated using the publicly available code and trained on the same training datasets for fair comparisons.}
\vspace{-0.5mm}
    \begin{tabular}{lcccc}
    \hline
    Methods                         &NIQE $\downarrow$          &CLIPIQA+ $\uparrow$           &MUSIQ-PAQ2PIQ $\uparrow$        \\ \hline
    MIMO-UNet~\cite{MIMO}           &9.9320                     &0.3130             &59.8464              \\
    NAFNet~\cite{NAFNet}            &9.2736                     &0.3045             &59.7057              \\
    Restormer~\cite{Restormer}      &9.6038                     &0.3051             &60.7811              \\
    MIMO-UNet*~\cite{MIMO}          &8.2208                     &0.2950             &63.6510              \\
    NAFNet*~\cite{NAFNet}           &8.2431                     &0.2977             &63.6596              \\
    Restormer*~\cite{Restormer}     &8.5662                     &0.3127             &64.7834              \\
    LBAG~\cite{ReLoBlur}            &10.0196                    &0.2968             &58.5612              \\
    Ours                            &\textbf{4.3228}            &\textbf{0.3531}    &\textbf{65.1461}     \\ \hline
\end{tabular}
\label{compare2}
\vspace{-4mm}
\end{table}
To provide a comprehensive evaluation, we also compare it with global deblurring methods such as NAFNet~\cite{NAFNet}, Restormer~\cite{Restormer}, and MIMO-UNet~\cite{MIMO} and local deblurring methods such as LBAG~\cite{ReLoBlur} and LMD-ViT~\cite{LMD-ViT}.

As the proposed method requires a short-exposure image as guidance, one may wonder whether the performance gains are due to the use of additional information. To answer this question, we further use the concatenation of the blurry and the short-exposure images as input of the existing methods for comparison.

For fair comparisons, we retrain the evaluated methods on the same training datasets.
The values of NIQE, CLIPIQA+, and MUSIQ-PAQ2PIQ are tested on images of size $512 \times 512$ with locally blurred.
\vspace{-2mm}

{\flushleft \textbf{Evaluations on the ReLoBlur synthetic dataset.}}
Table~\ref{compare1} summarizes the quantitative evaluation results, where the proposed method achieves high-quality results with higher NIQE, CLIPIQA+, and MUSIQ-PAQ2PIQ values.

\begin{figure*}[!t]
\footnotesize
\centering
    \begin{tabular}{c c c c c c c}
            \multicolumn{3}{c}{\multirow{5}*[45.6pt]{
            \hspace{-2.5mm} \includegraphics[width=0.325\linewidth,height=0.235\linewidth]{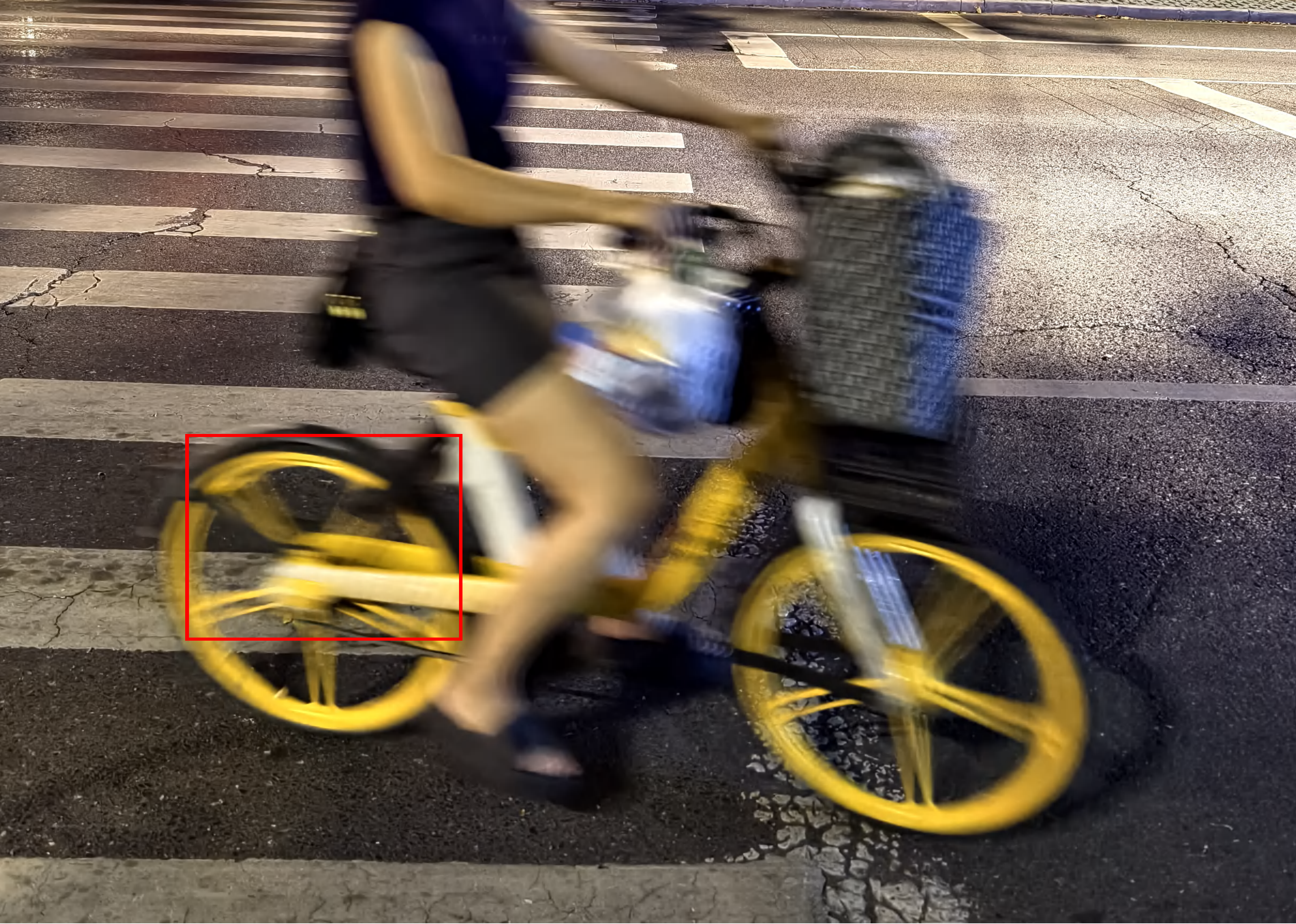}}}
            & \hspace{-4.0mm} \includegraphics[width=0.16\linewidth,height=0.105\linewidth]{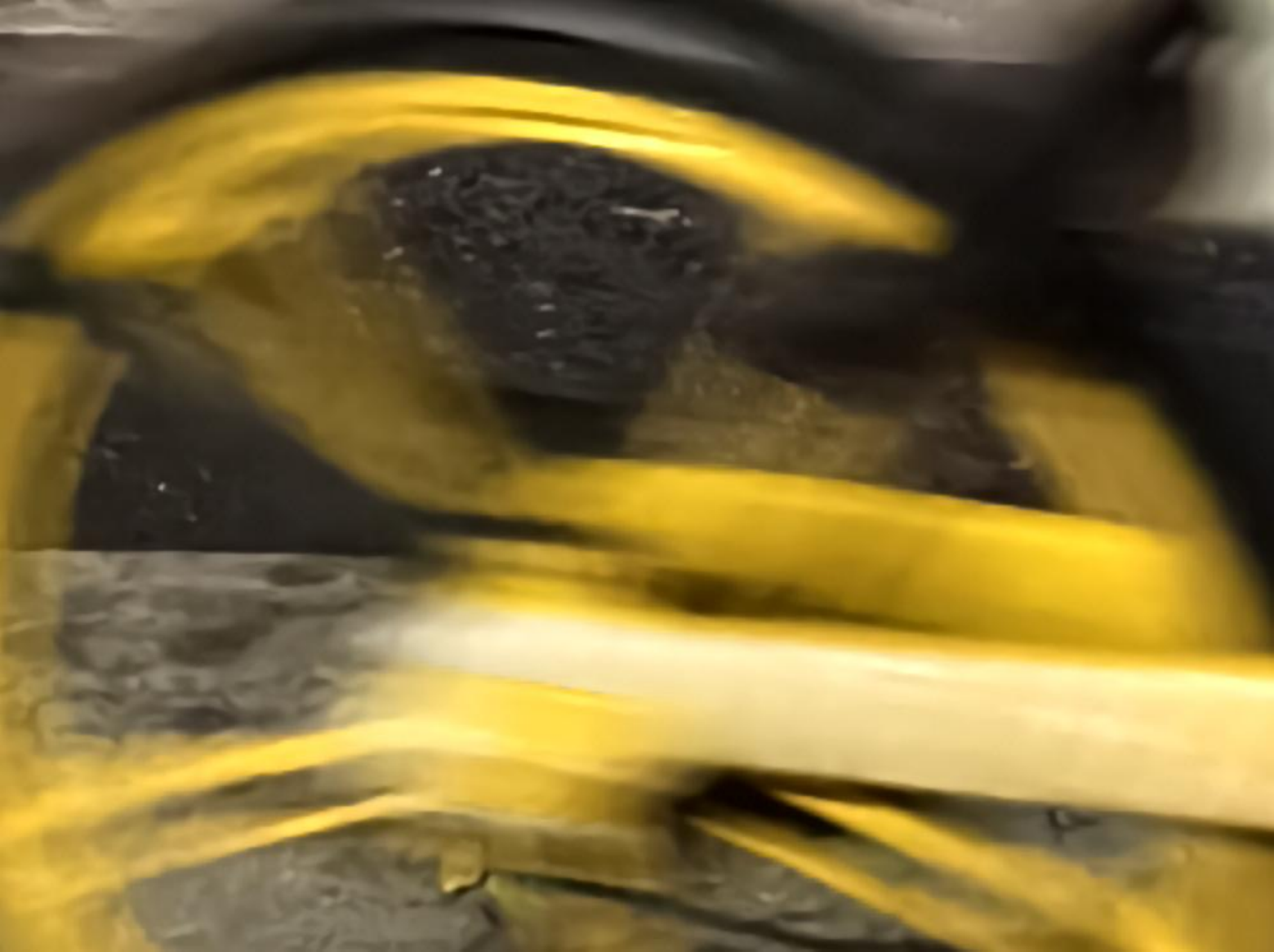}
            & \hspace{-4.0mm} \includegraphics[width=0.16\linewidth,height=0.105\linewidth]{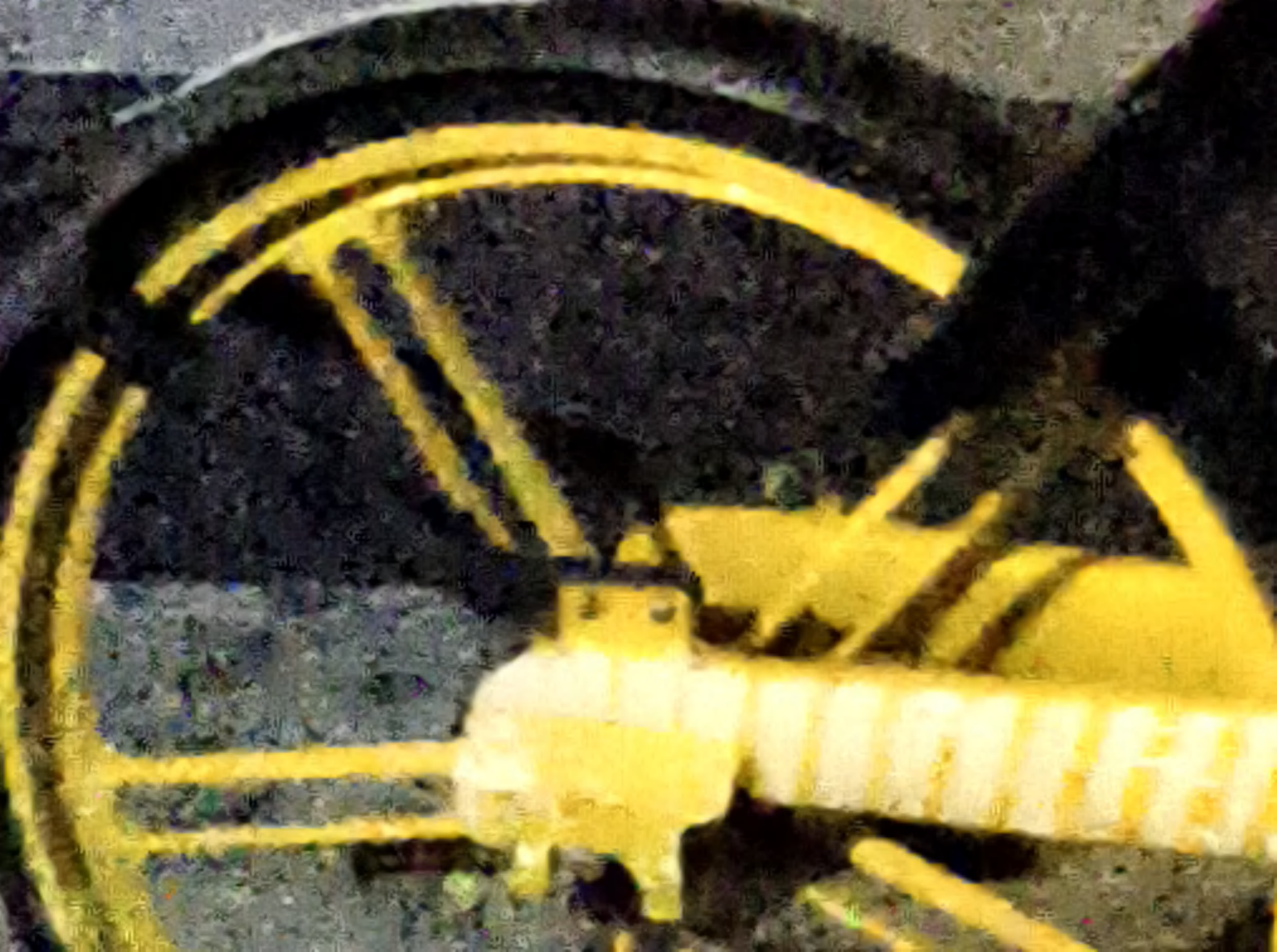}
            & \hspace{-4.0mm} \includegraphics[width=0.16\linewidth,height=0.105\linewidth]{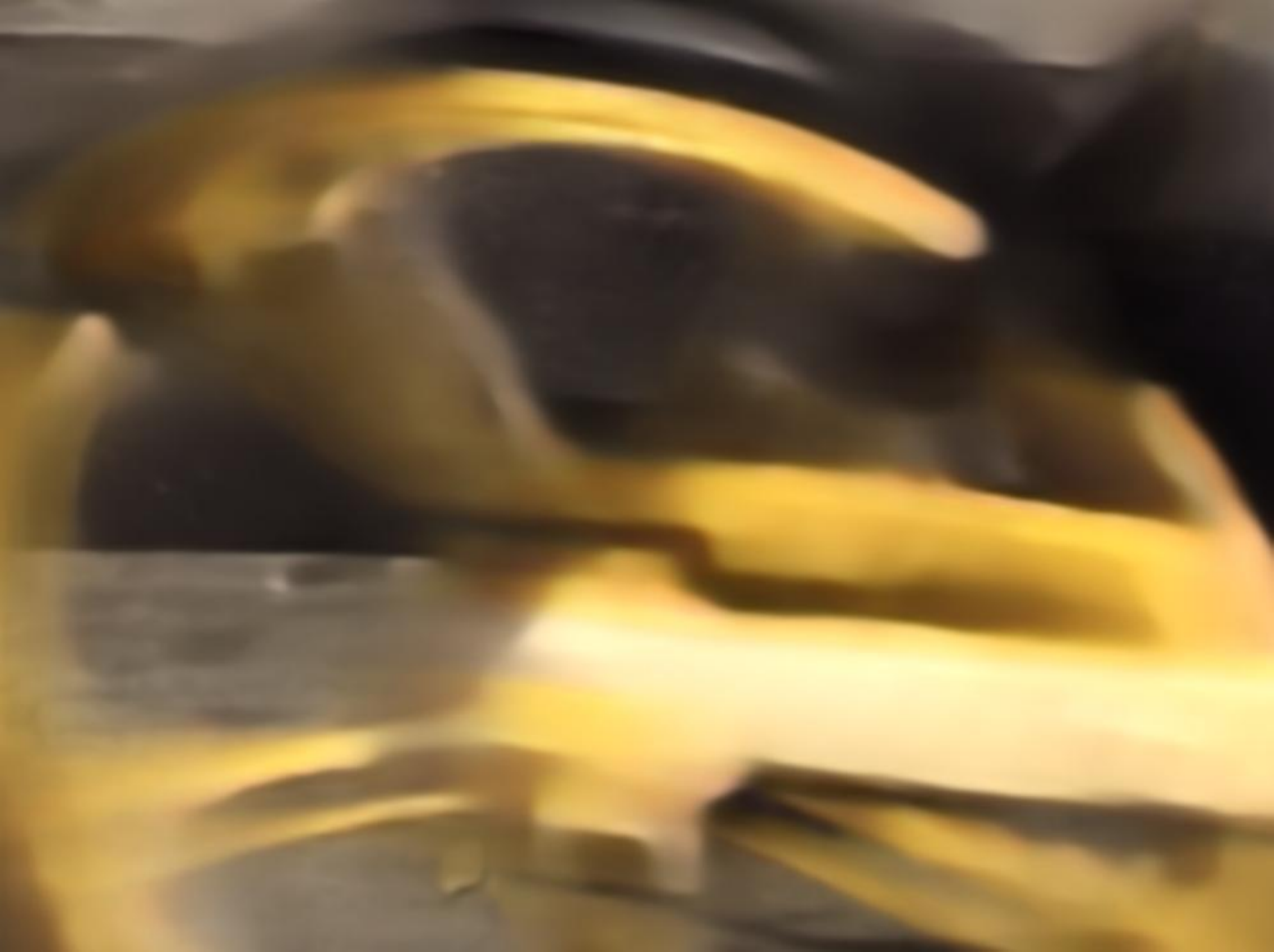}
            & \hspace{-4.0mm} \includegraphics[width=0.16\linewidth,height=0.105\linewidth]{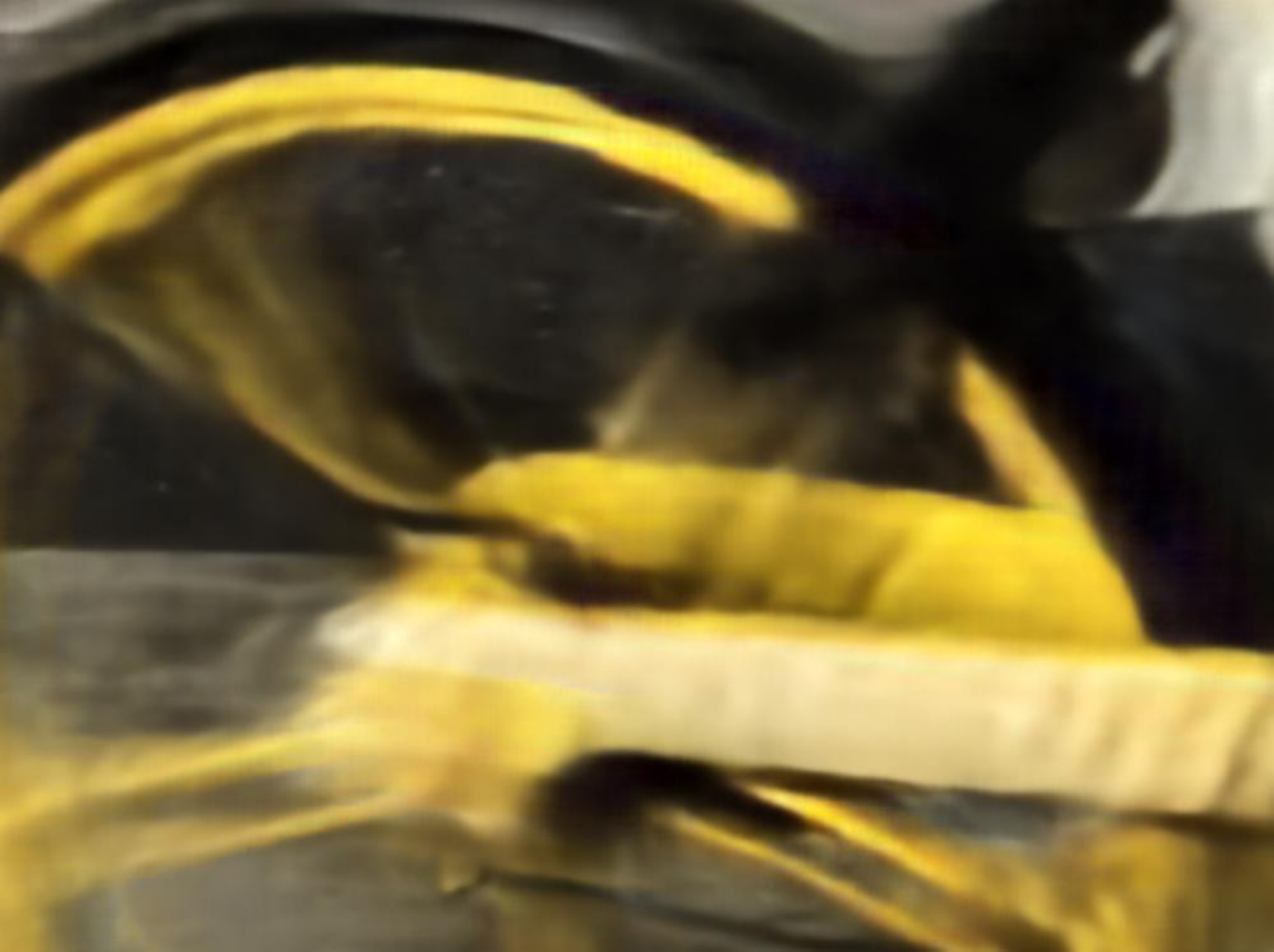}

              \\
    		\multicolumn{3}{c}{~}
            & \hspace{-4.0mm} (a) Blurred patch
            & \hspace{-4.0mm} (b) Short-exposure patch
            & \hspace{-4.0mm} (c) LBAG~\cite{ReLoBlur}
            & \hspace{-4.0mm} (d) NAFNet~\cite{NAFNet}
\\	

    	\multicolumn{3}{c}{~}
            & \hspace{-4.0mm} \includegraphics[width=0.16\linewidth,height=0.105\linewidth]{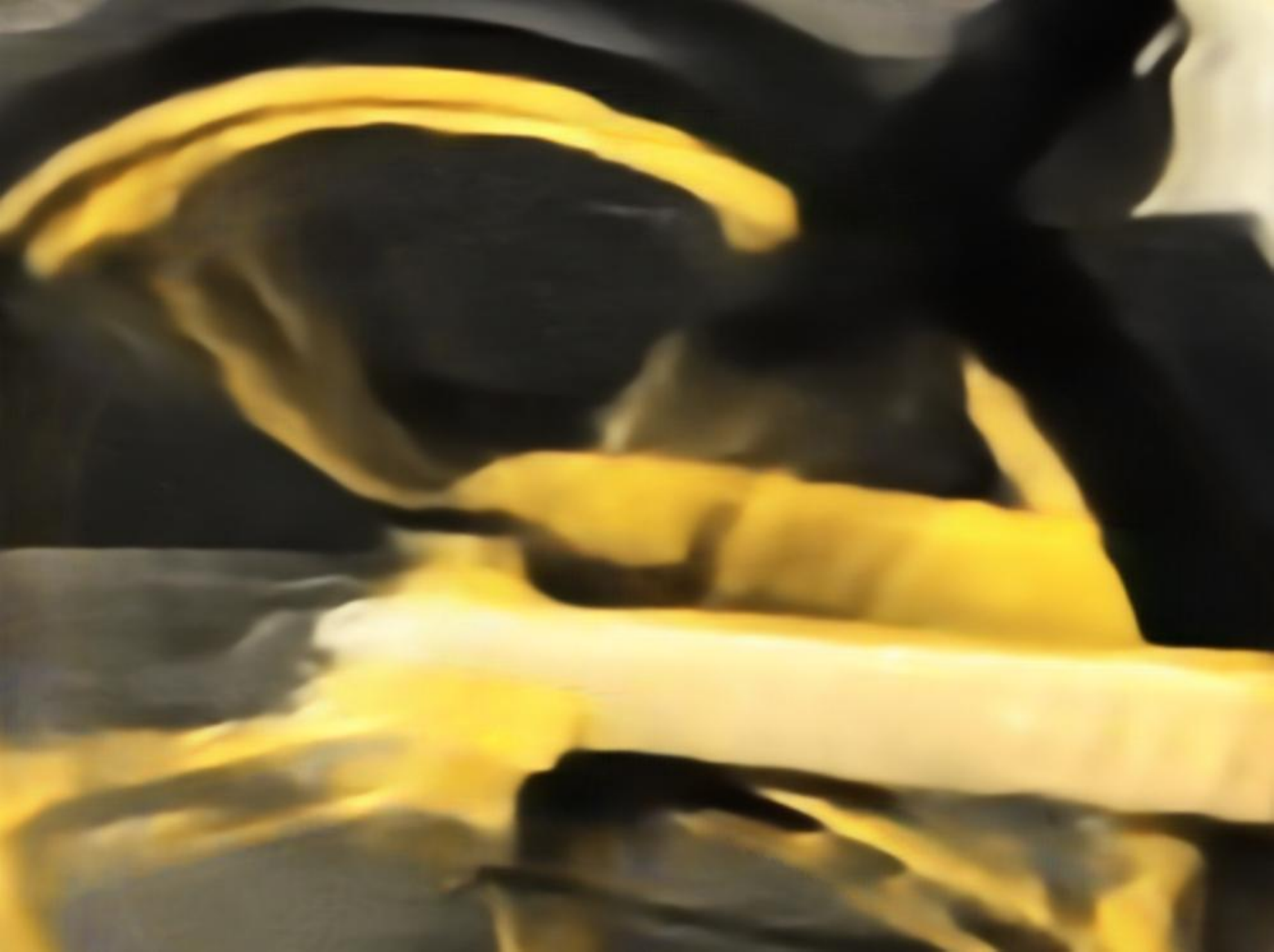}
            & \hspace{-4.0mm} \includegraphics[width=0.16\linewidth,height=0.105\linewidth]{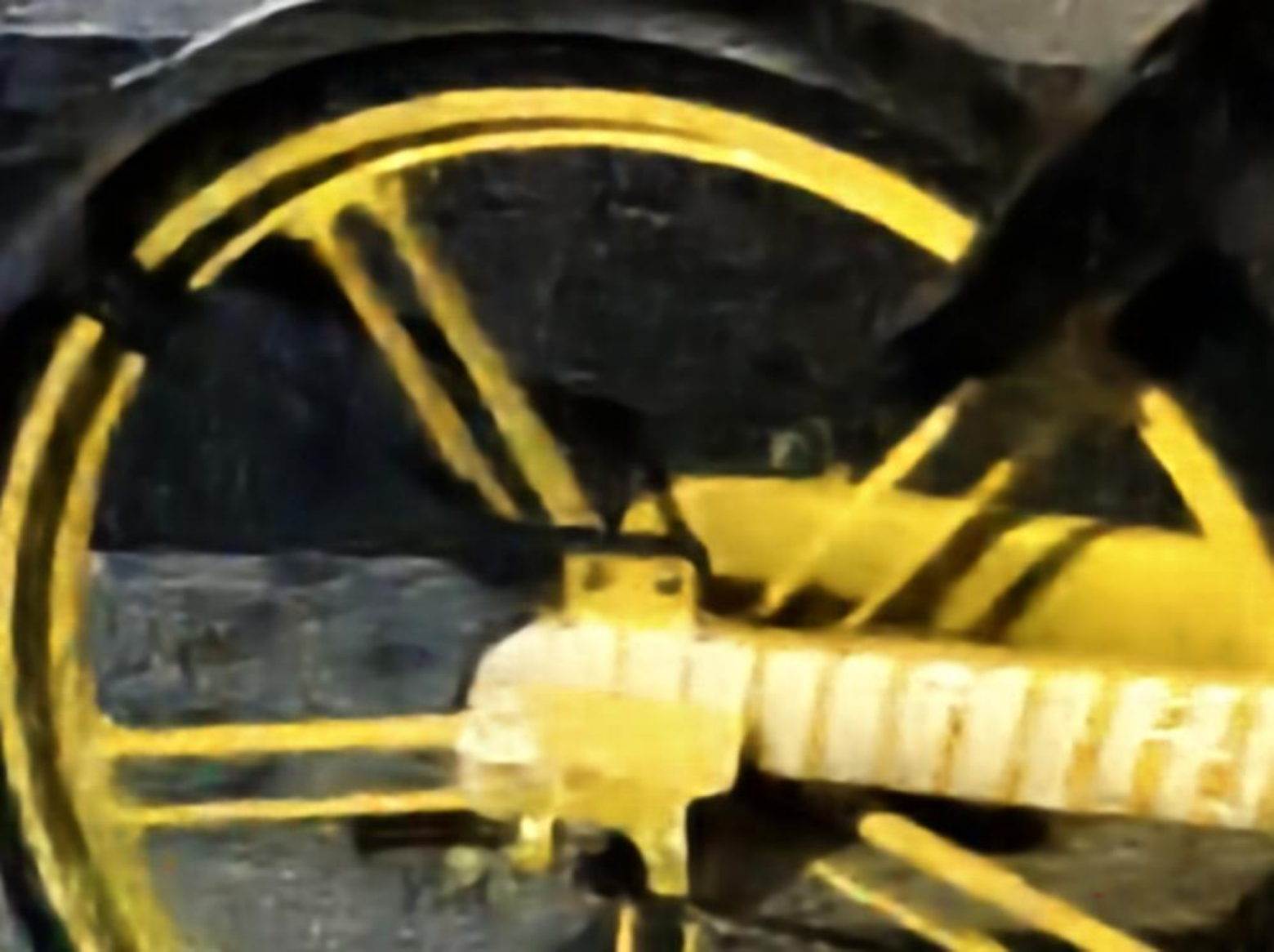}
            & \hspace{-4.0mm} \includegraphics[width=0.16\linewidth,height=0.105\linewidth]{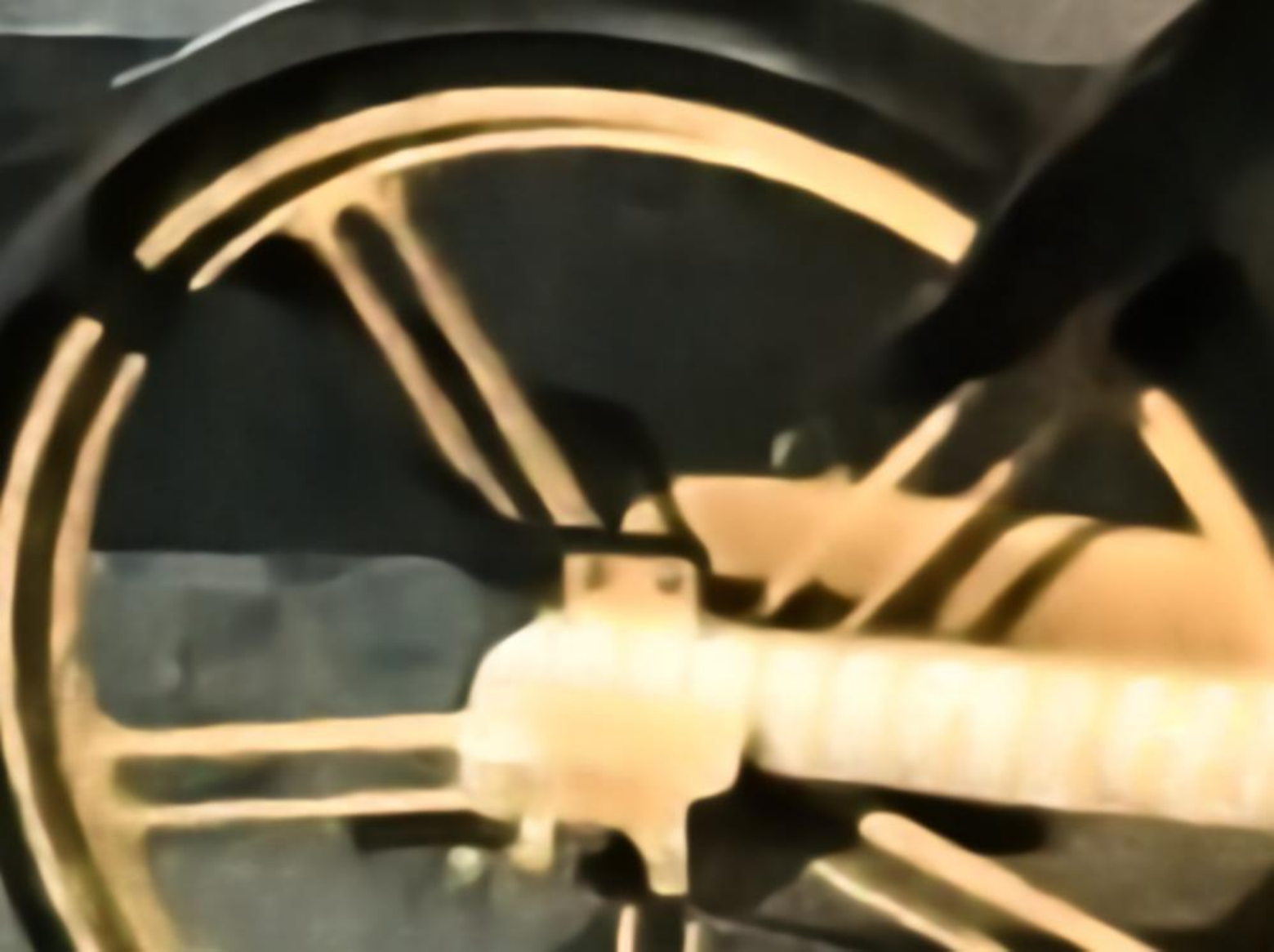}
            & \hspace{-4.0mm} \includegraphics[width=0.16\linewidth,height=0.105\linewidth]{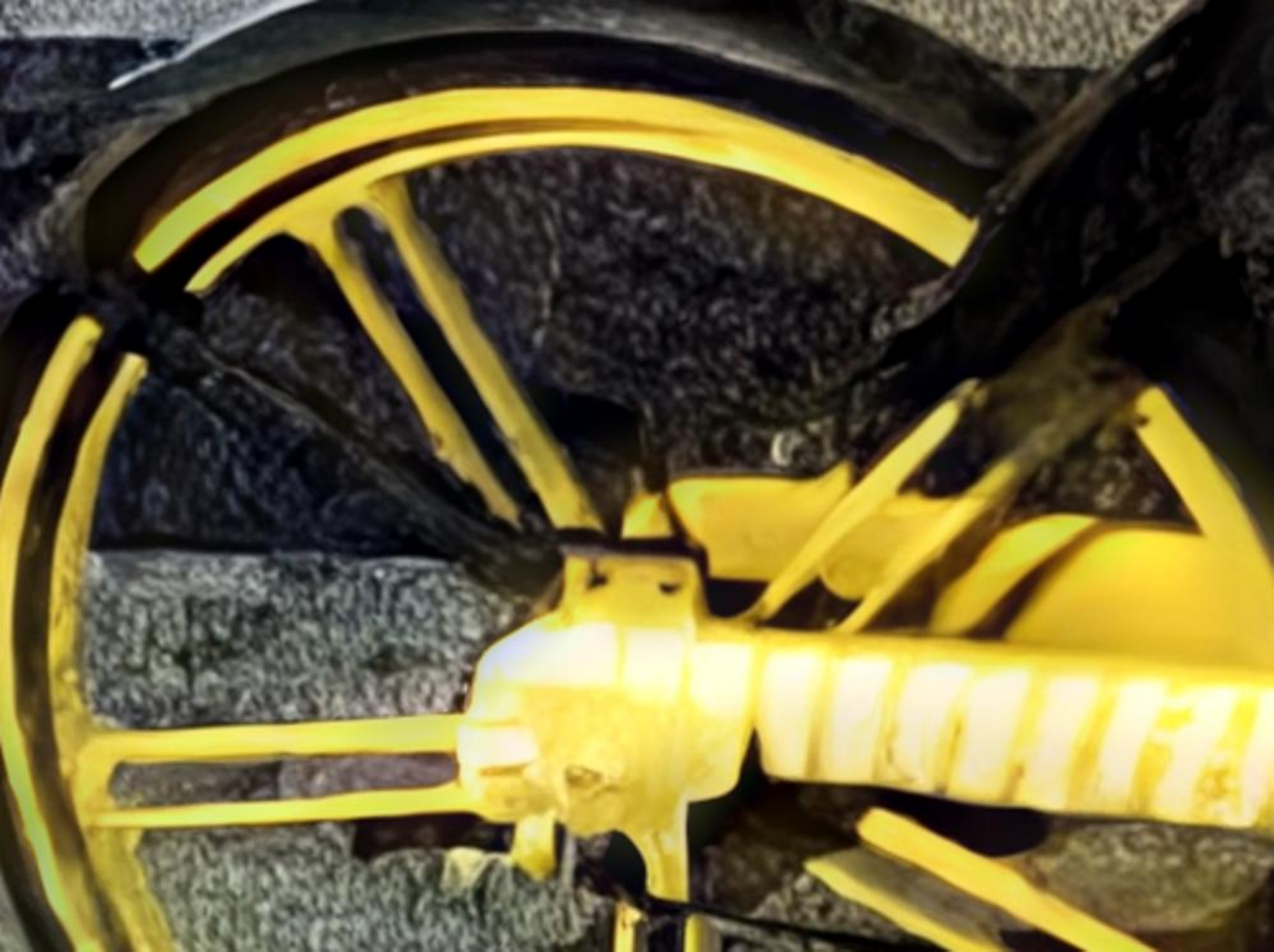}
            \\

    	\multicolumn{3}{c}{\hspace{-4.0mm} \textbf{Blurry input}}
            & \hspace{-4.0mm} (e) Restormer~\cite{Restormer}
            & \hspace{-4.0mm} (f) NAFNet*~\cite{NAFNet}
            & \hspace{-4.0mm} (g) Restormer*~\cite{Restormer}
            & \hspace{-4.0mm} (h) Ours
            \\
            \multicolumn{3}{c}{\multirow{5}*[45.6pt]{
            \hspace{-2.5mm} \includegraphics[width=0.325\linewidth,height=0.235\linewidth]{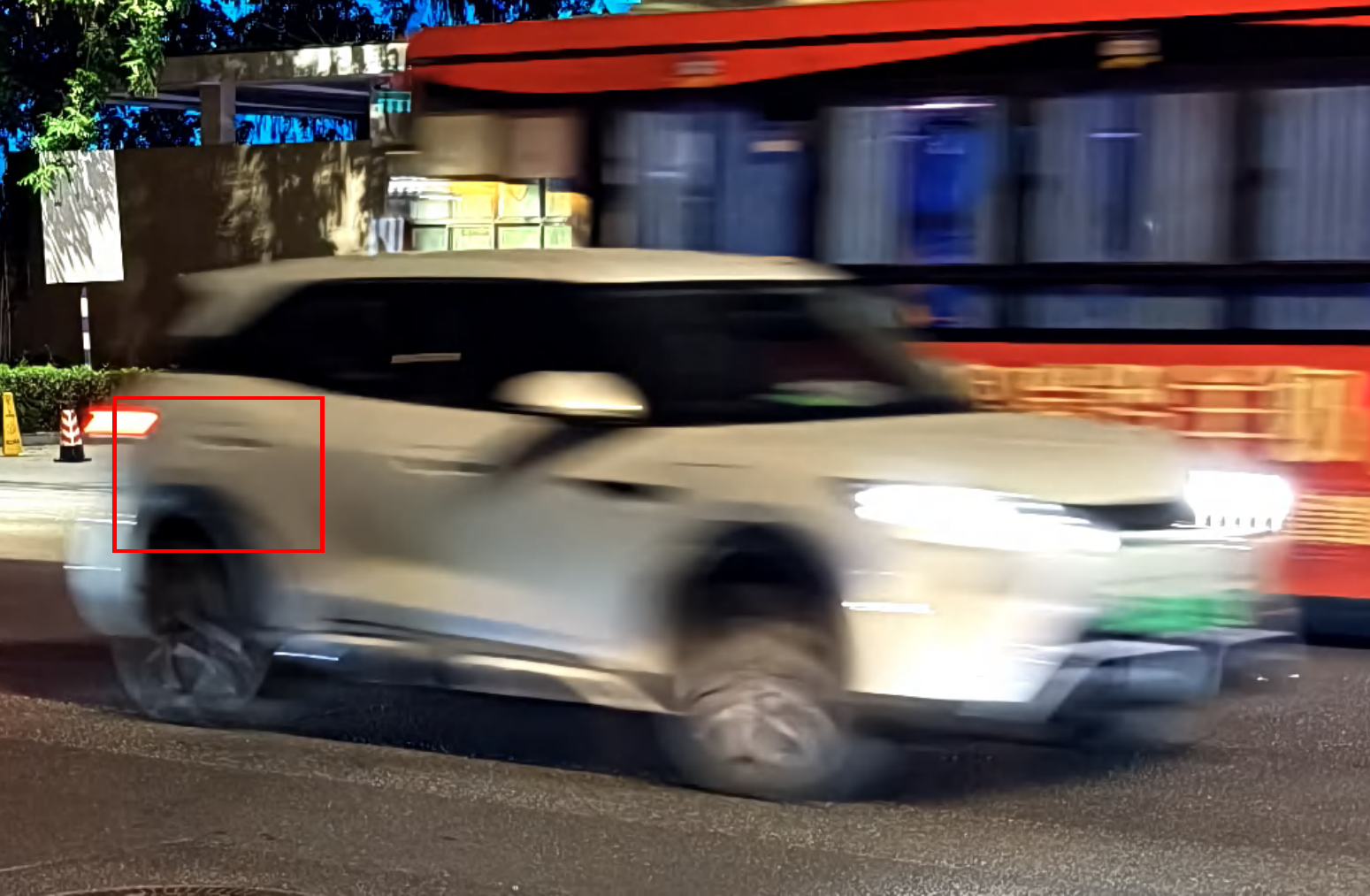}}}
            & \hspace{-4.0mm} \includegraphics[width=0.16\linewidth,height=0.105\linewidth]{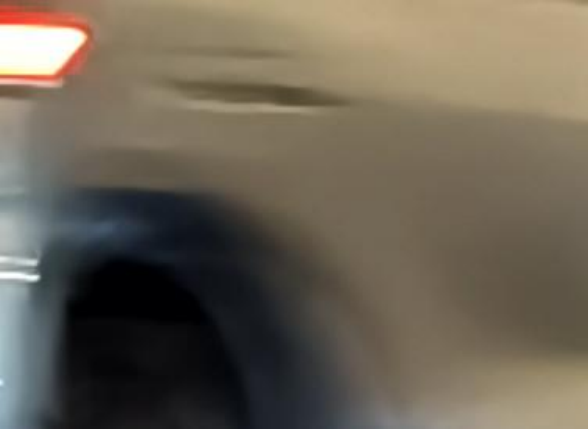}
            & \hspace{-4.0mm} \includegraphics[width=0.16\linewidth,height=0.105\linewidth]{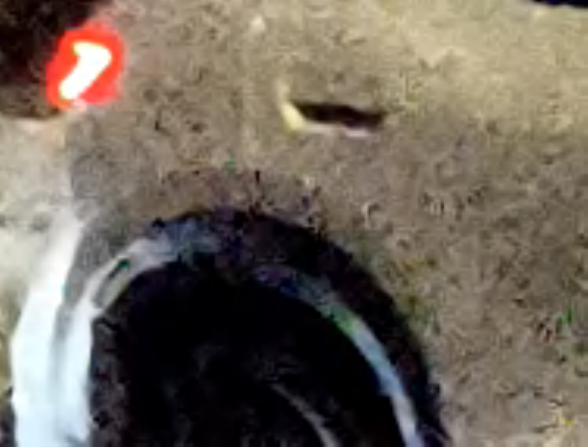}
            & \hspace{-4.0mm} \includegraphics[width=0.16\linewidth,height=0.105\linewidth]{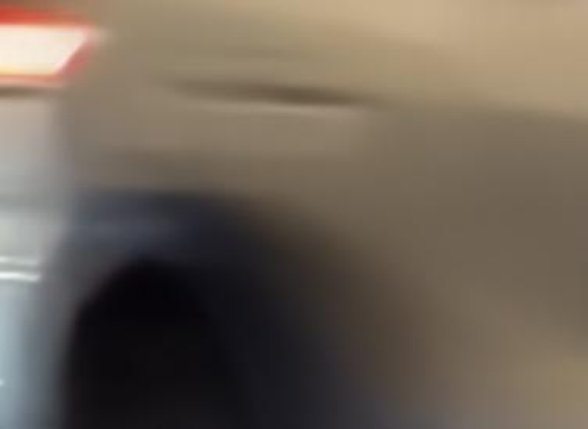}
            & \hspace{-4.0mm} \includegraphics[width=0.16\linewidth,height=0.105\linewidth]{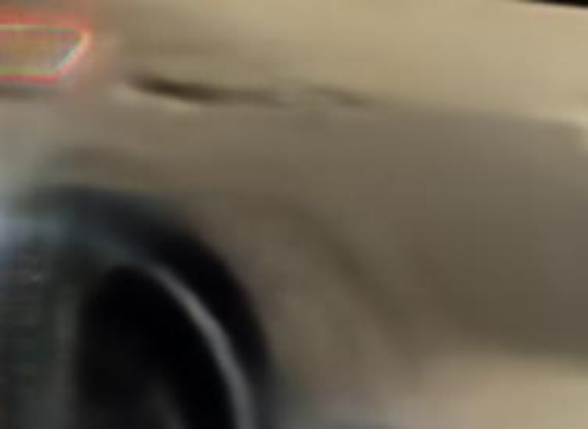}

              \\
    		\multicolumn{3}{c}{~}
            & \hspace{-4.0mm} (a) Blurred patch
            & \hspace{-4.0mm} (b) Short-exposure patch
            & \hspace{-4.0mm} (c) LBAG~\cite{ReLoBlur}
            & \hspace{-4.0mm} (d) NAFNet~\cite{NAFNet}
\\	

    	\multicolumn{3}{c}{~}
            & \hspace{-4.0mm} \includegraphics[width=0.16\linewidth,height=0.105\linewidth]{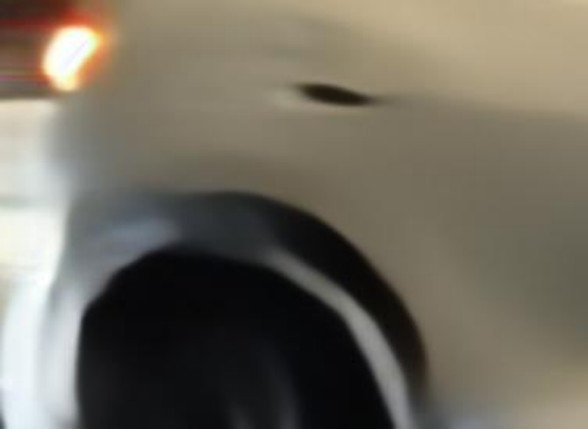}
            & \hspace{-4.0mm} \includegraphics[width=0.16\linewidth,height=0.105\linewidth]{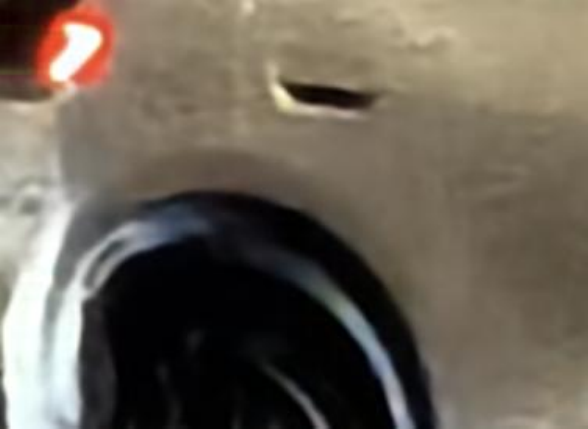}
            & \hspace{-4.0mm} \includegraphics[width=0.16\linewidth,height=0.105\linewidth]{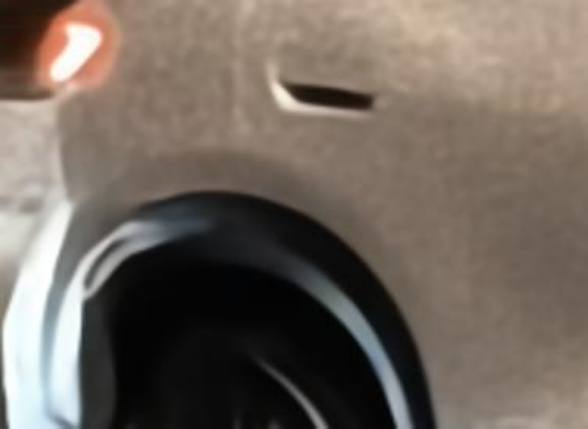}
            & \hspace{-4.0mm} \includegraphics[width=0.16\linewidth,height=0.105\linewidth]{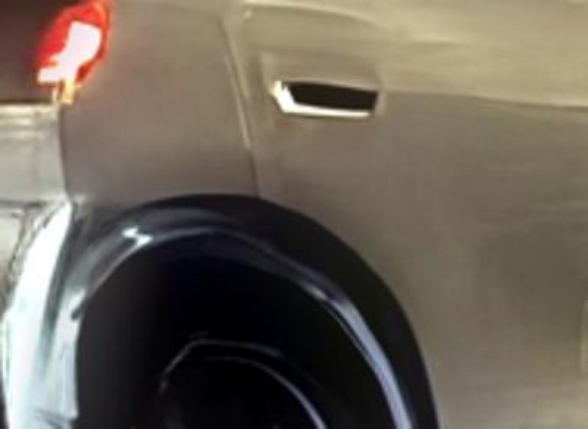}
            \\

    	\multicolumn{3}{c}{\hspace{-4.0mm} \textbf{Blurry input}}
            & \hspace{-4.0mm} (e) Restormer~\cite{Restormer}
            & \hspace{-4.0mm} (f) NAFNet*~\cite{NAFNet}
            & \hspace{-4.0mm} (g) Restormer*~\cite{Restormer}
            & \hspace{-4.0mm} (h) Ours
            \\

    \end{tabular}
\caption{Deblurred results on the LMBD dataset. The deblurred results in (c)-(g) still contain significant blur effects. The proposed method generates a better deblurred image with clearer structures in (h).}
\label{fig:visual2}
\vspace{-6mm}
\end{figure*}
Although the global deblurring methods use short-exposure images as reference (e.g., Restormer*~\cite{Restormer}), they show improvement in the results compared to when no reference image guidance, the NIQE gain of our method is at least 0.62 better than above methods.
In addition, compared to the local deblurring methods~\cite{ReLoBlur, LMD-ViT}, our method generates better results.
Figure~\ref{fig:visual1} shows that the evaluated methods do not restore the shoes effectively. In contrast, our method generates better results.
For example, the shoelaces and boundaries are much clearer.
%

%
\begin{table}[]\footnotesize
\caption{Effect of the proposed blurry-aware guided image restoration in the proposed method on the LMBD dataset.}
\vspace{-2mm}
\label{tab:abation1}
\begin{tabular}{lccc}
\hline
Methods                           &~~~~${\text{w/o BGF}}$~~~~   &~~~~~${\text{w/ GF}}$~~~~     &~~~~$\text{w/ BGF}$~~~~      \\  \hline
CLIPIQA+ $\uparrow$                & 0.3296                  & 0.2815                    & \textbf{0.3395}          \\
MUSIQ-PAQ2PIQ $\uparrow$           & 62.1051                 & 59.5076                   & \textbf{62.4996}                  \\  \hline
\end{tabular}
\vspace{-1.2mm}
\end{table}
\begin{figure}[]
\footnotesize
\centering
    \begin{tabular}{cccc}
    \hspace{-2mm} \includegraphics[width=0.11\textwidth]{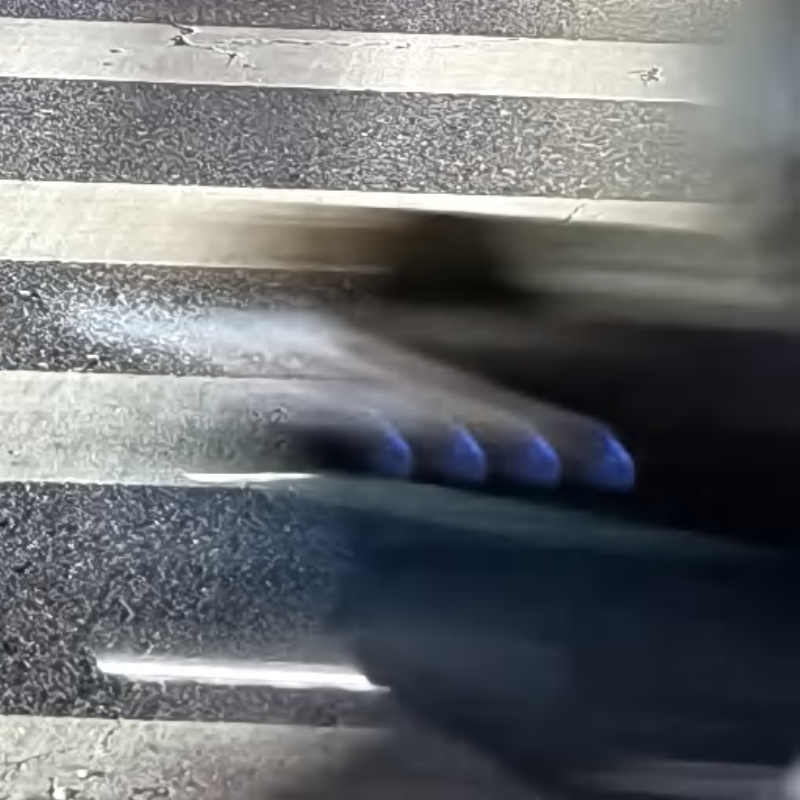}
    & \hspace{-3mm} \includegraphics[width=0.11\textwidth]{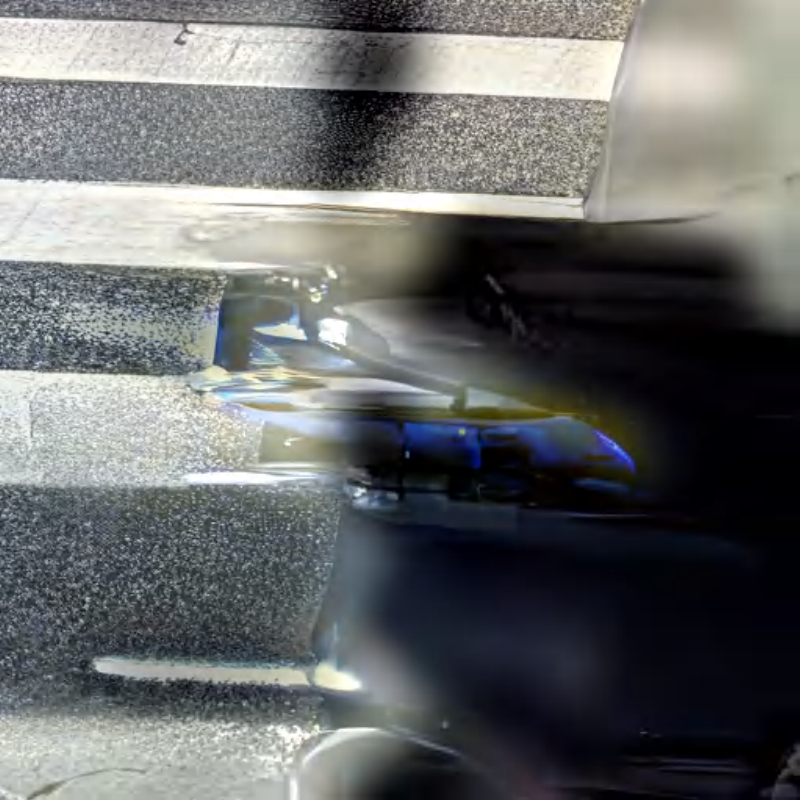}
    & \hspace{-3mm} \includegraphics[width=0.11\textwidth]{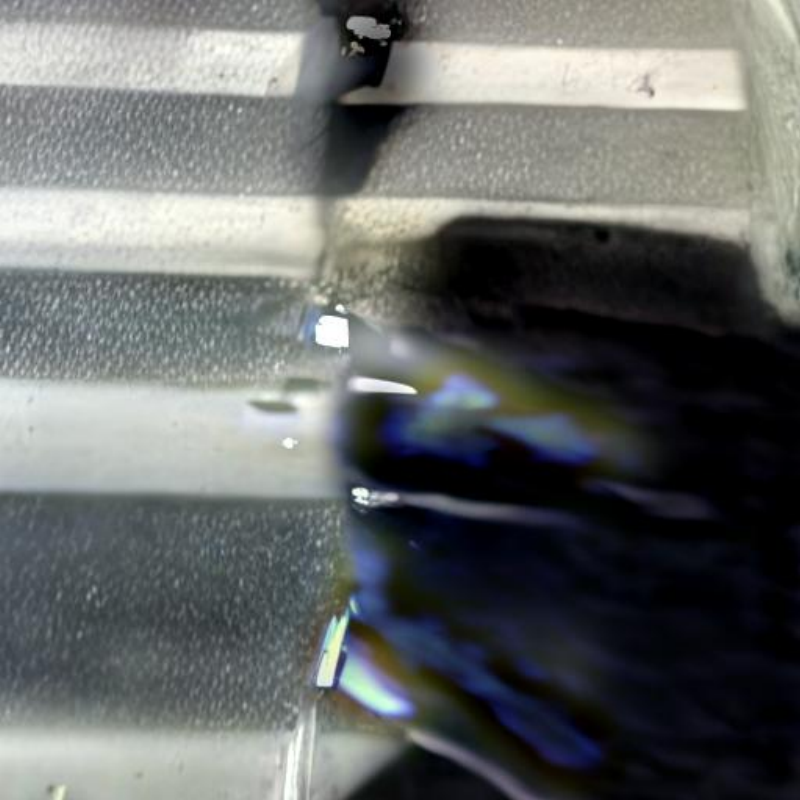}
    & \hspace{-3mm} \includegraphics[width=0.11\textwidth]{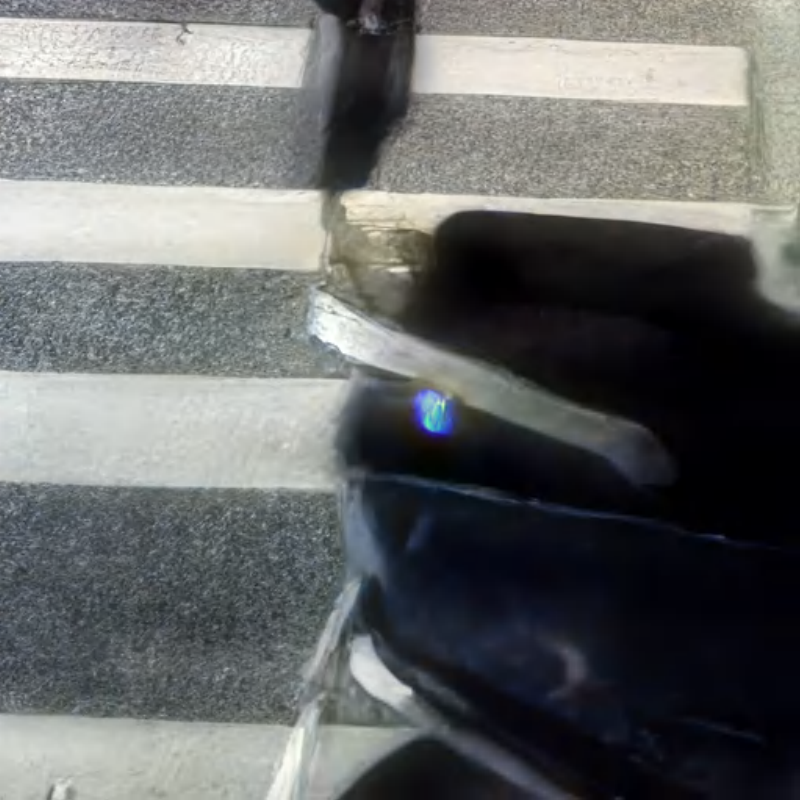} \\
    \hspace{-2mm}(a) Blurry input &\hspace{-3.7mm} (b) ${\text{w/o BGF}}$ &\hspace{-5.1mm} (c) ${\text{w/ GF}}$ &\hspace{-5.2mm} (d) ${\text{w/ BGF}}$\\
    \end{tabular}
\vspace{-2mm}
    \caption{Effectiveness of the proposed blurry-aware guided image restoration on image deblurring.}
    \label{fig:ablation1}
\vspace{-5mm}
\end{figure}

{\flushleft \textbf{Evaluations on the LMBD dataset.}}
Table~\ref{compare2} shows the quantitative evaluation results.
Our method achieves the highest quantitative results.
Compared with global deblurring methods, our method shows a NIQE gain of at least 4.6 better than NAFNet~\cite{NAFNet}.
The above methods use the short-exposure image as the reference, which demonstrate measurable enhancements in their quantitative performance metrics compared to when reference image guidance is not provided.
However, the MUSIQ-PAQ2PIQ gain of our method is at least 0.52 higher than Restormer*~\cite{NAFNet} and 6.74 higher than local deblurring method LBAG~\cite{ReLoBlur}.
\begin{table}[]\footnotesize
\caption{Effect of proposed ExpBFusion in the proposed method on the LMBD dataset.}
\vspace{-2mm}
\label{tab:abation2}
\begin{tabular}{lccc}
\hline
Methods                  &~~~${\text{w/o F}}$~~~       &~~~~${\text{w/ F \& w/o~} M}$~~~~~       &~~~${\text{w/ F}}$~~~      \\ \hline
CLIPIQA+ $\uparrow$                 & 0.3395                 & 0.3368                       & \textbf{0.3531}             \\
MUSIQ-PAQ2PIQ $\uparrow$            & 62.4996                & 64.7102                      & \textbf{65.1461}            \\ \hline
\end{tabular}
\vspace{-1.7mm}
\end{table}
\begin{figure}[]
\footnotesize
\centering
    \begin{tabular}{cccc}
    \hspace{-2mm} \includegraphics[width=0.11\textwidth]{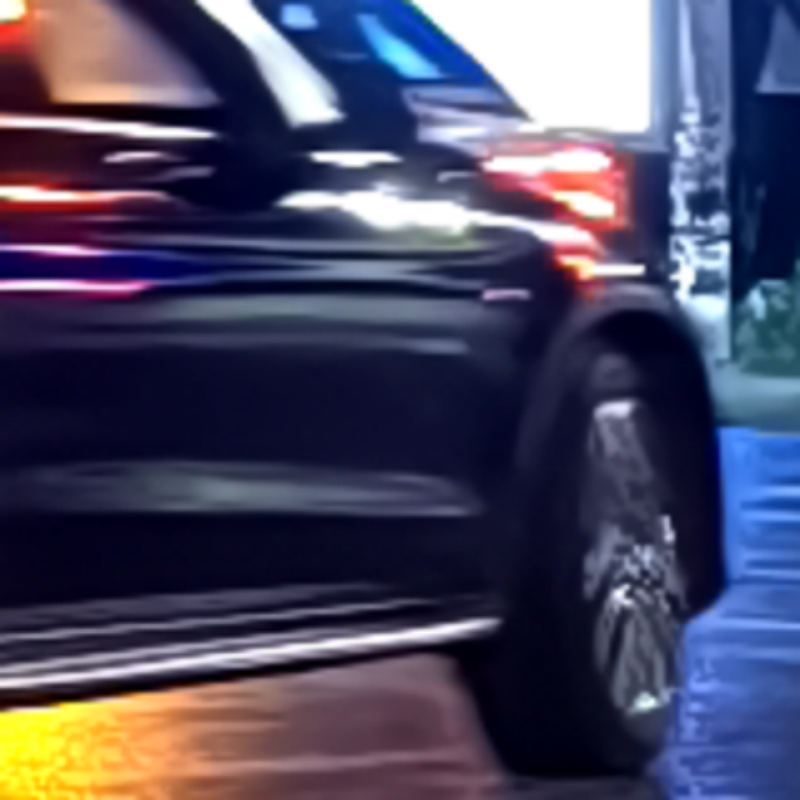}
    & \hspace{-3mm} \includegraphics[width=0.11\textwidth]{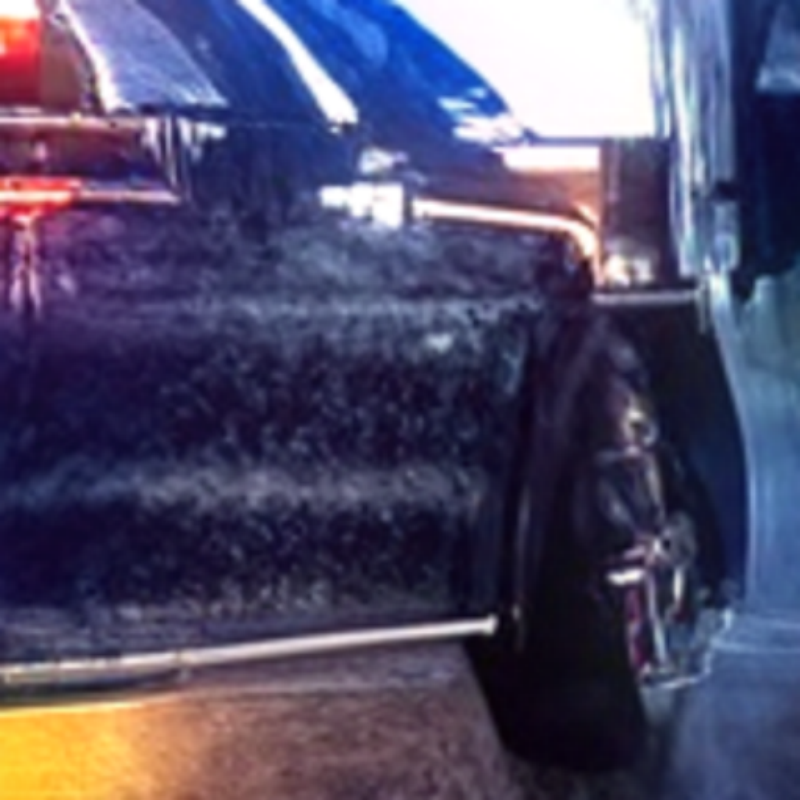}
    & \hspace{-3mm} \includegraphics[width=0.11\textwidth]{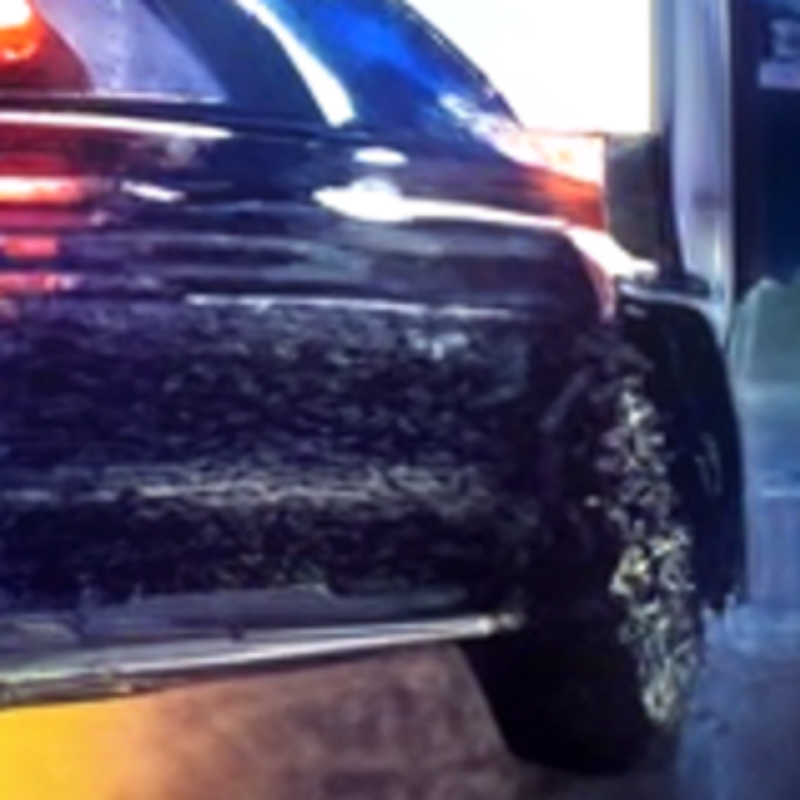}
    & \hspace{-3mm} \includegraphics[width=0.11\textwidth]{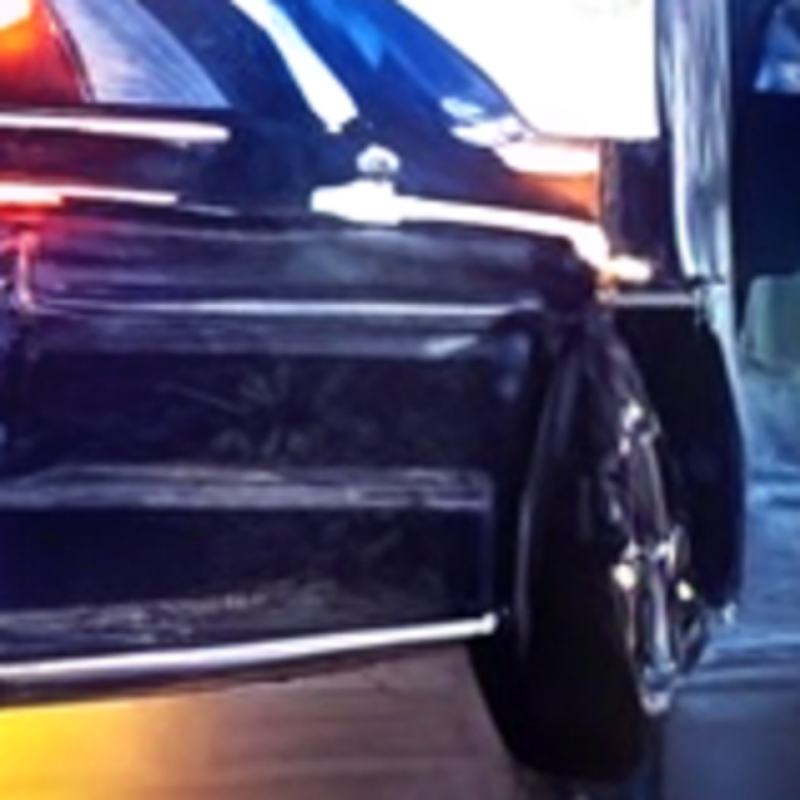} \\
    \hspace{-2mm}(a) Blurry input &\hspace{-3.7mm} (b) ${\text{w/o F}}$ &\hspace{-5.1mm} (c) ${\text{w/ F \& w/o~} M}$ &\hspace{-5.2mm} (d) ${\text{w/ F}}$\\
    \end{tabular}
\vspace{-2mm}
    \caption{Effectiveness of the proposed ExpBFusion on image deblurring.}
    \label{fig:ablation2}
\vspace{-5mm}
\end{figure}
Figure~\ref{fig:visual2} shows some visual comparisons of the evaluated methods.
The global deblurring methods do not remove the blur effects, as shown in Figure~\ref{fig:visual2}(c)-(d).
Figure~\ref{fig:visual2}(f)-(g) are the visual results of the above methods that use the short-exposure image as a reference and are still missing structural information.
In contrast to the above methods, we utilize the priors from diffusion models to assist in the restoration of blurred regions to achieve a refinement effect in the blurred regions.
Thus, the deblurred results contain clear structures, and the hub of the tire and the ground appear much clearer as shown in Figure~\ref{fig:visual2}(h).
\vspace{-3mm}

\section{Analysis and Discussion}
\label{ablation}
To better understand how our method addresses local motion deblurring with reference images,
and demonstrate the effect of its main components, we provide in-depth analysis of the proposed method.
For the ablation studies in this section, we train our method and all the baselines on the LMBD dataset using the batch size of 32.
{\flushleft \textbf{Effect of the blurry-aware guided image restoration.}}
The proposed blurry-aware guided image restoration is designed to eliminate blur with the guidance of a short-exposure image in blurred regions.

To demonstrate the effectiveness of our proposed blur-aware guided image restoration, we compare the method utilizing blurry-aware guided image restoration (${\text{w/ BGF}}$ for short) with the baseline that the method uses long-exposure image (${\text{w/o BGF}}$ for short) to guide the restoration of the local motion blurred image.

Table~\ref{tab:abation1} shows that the method using the blur-aware guided image restoration generates better deblurring results, where the MUSIQ-PAQ2PIQ value is 0.39 better.
Furthermore, we compare the method using the blur-aware guided image restoration with the method using guided filtering method~\cite{GIF} (${\text{w/ GF}}$ for short).
The comparison results table~\ref{tab:abation1} shows that the method using blur-aware guided image restoration generates better results with higher MUSIQ-PAQ2PIQ values.
Figure~\ref{fig:ablation1} shows that the method using the proposed blur-aware guided image restoration generates better deblurred images, where the motorcycle tail is recovered well shown in Figure~\ref{fig:ablation1}(d).

\begin{figure}[t]
\footnotesize
\centering
    \begin{tabular}{cccc}
    \hspace{-2mm} \includegraphics[width=0.11\textwidth]{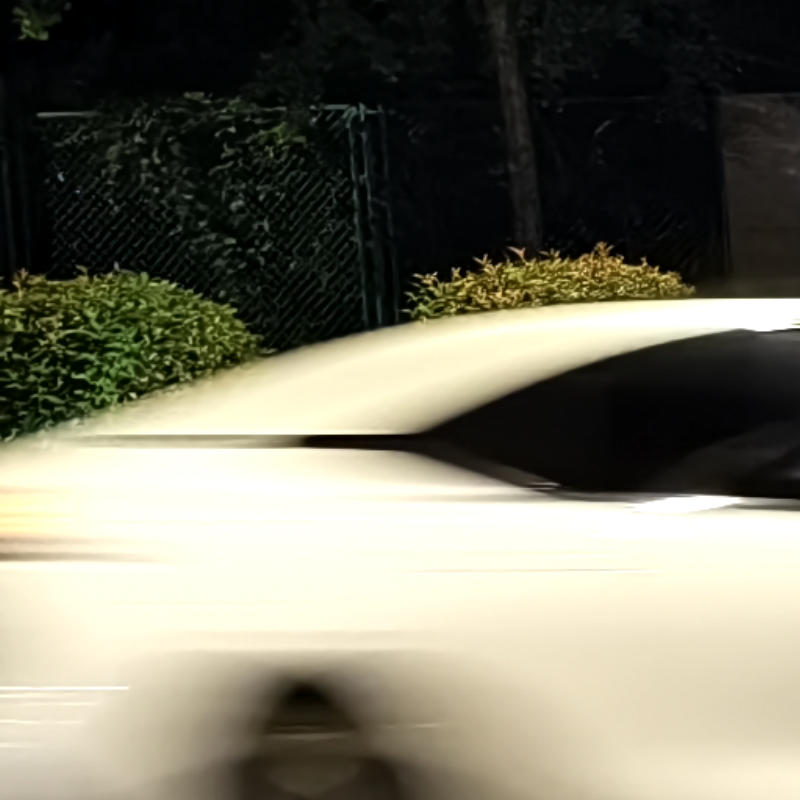}
    & \hspace{-3mm} \includegraphics[width=0.11\textwidth]{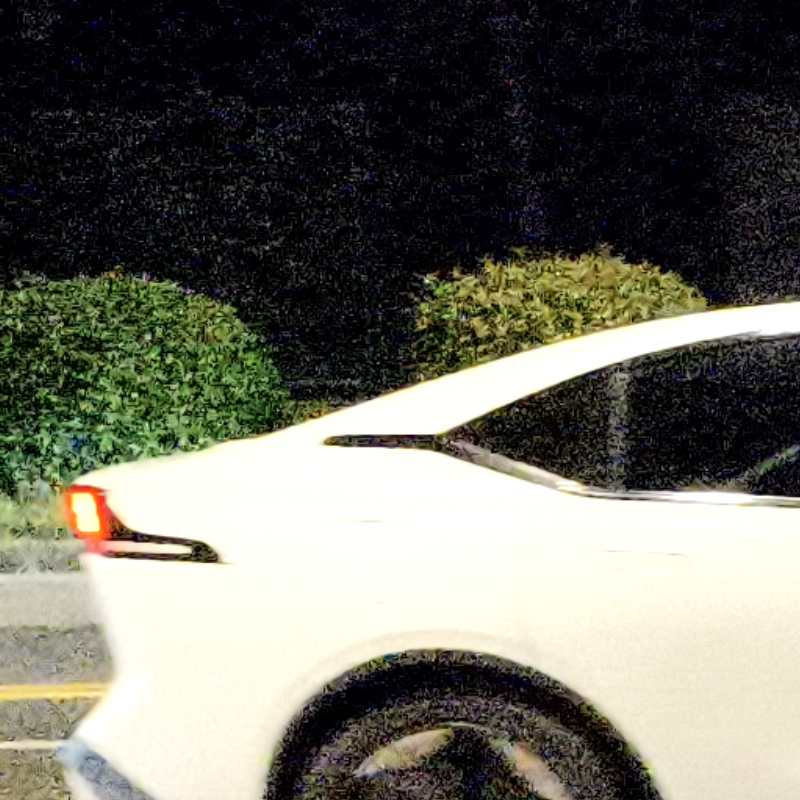}
    & \hspace{-3mm} \includegraphics[width=0.11\textwidth]{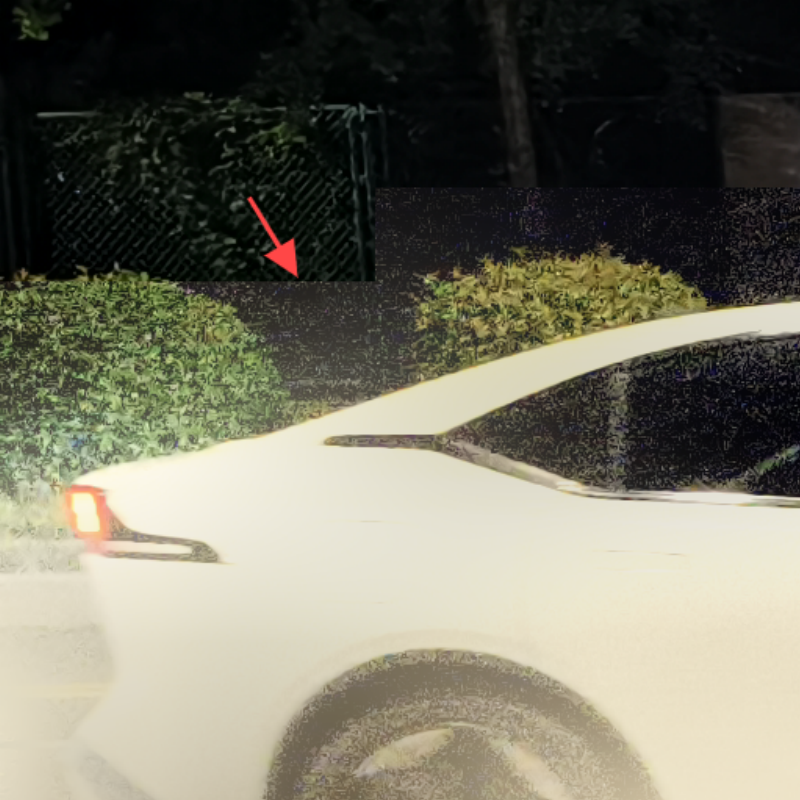}
    & \hspace{-3mm} \includegraphics[width=0.11\textwidth]{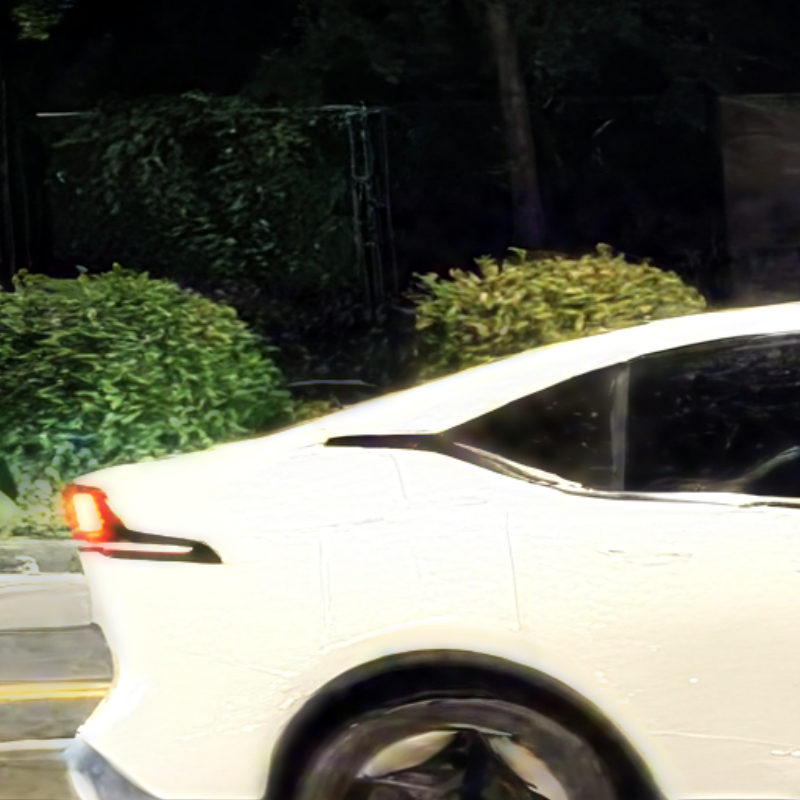} \\
    \hspace{-2mm}(a)  &\hspace{-3.7mm} (b)  &\hspace{-5.1mm} (c)  &\hspace{-5.2mm} (d) \\
    \end{tabular}
    \vspace{-1mm}
    \caption{Effectiveness of the proposed short-exposure guided diffusion model on the elimination of boundary artifacts. (a) Blur input; (b) short-exposure reference; (c) result of blurry-aware guided image restoration; (d) result of our method.}

    \label{fig:arxiv1}
    \vspace{-5mm}
\end{figure}

\label{ablation52}
{\flushleft \textbf{Effect of the short-exposure guided diffusion model.}}
The proposed short-exposure guided diffusion model is used to generate realistic results by a fusion module ExpBFusion.
To examine whether ExpBFusion effectively utilizes features from both short-exposure images and blurred regions to guide the diffusion process, we compare our method (${\text{w/ F}}$ for short) with the baseline that the proposed method without using a short-exposure guided diffusion model (${\text{w/o F}}$ for short).
Table~\ref{tab:abation2} shows that the proposed method using ExpBFusion generates better results, where the MUSIQ-PAQ2PIQ values is 2.80 higher than baseline.
Furthermore, we explore whether utilizing the global sharp information from the short-exposure image to the diffusion process improves the results.
To answer this question, we remove the $M$ from the ExpBFusion (${\text{w/ F \& w/o $M$}}$ for short) as the baseline.
Compared with the baseline, our proposed method generates better results, where the MUSIQ-PAQ2PIQ value is 0.44 better.
Figure~\ref{fig:ablation2} shows the visualization results.
Using the proposed short-exposure guided diffusion model generates better clear images, where the lines of the tires and the ground in the background are clearer.
We further explore the effectiveness of the proposed short-exposure guided diffusion model in handling the boundary effects caused by the mask. Figure~\ref{fig:arxiv1}(c) shows that after blurry-aware guided image restoration, although the blur has been removed, boundary effects introduced by the mask remain. However, after applying the short-exposure guided diffusion model, the result of our method eliminates the influence of these boundary effects, as shown in Figure~\ref{fig:arxiv1}(d).
\vspace{-1mm}

\begin{table}[]\footnotesize
\caption{Effect of proposed context-based local blur detection module in the proposed method on the LMBD dataset.}
\vspace{-5mm}
\label{tab:abation3}
\begin{center}
\begin{tabular}{l @{\hspace{4mm}} c @{\hspace{4mm}} c @{\hspace{4mm}} c}
\hline
Methods                                        &~~~~~${\text{w/~} M^{c}}$~~~~~         &~~~~~${\text{w/~}M}$~~~~~   &~~~~~${\text{w/~} M^{gt}}$~~~~  \\ \hline
CLIPIQA+ $\uparrow$                            & 0.3393                                & 0.3531                     & 0.3571                  \\
MUSIQ-PAQ2PIQ $\uparrow$                       & 64.4785                               & 65.1461                    & 65.3015                 \\ \hline
\end{tabular}
\end{center}
\vspace{-5mm}
\end{table}

\begin{figure}[]
\footnotesize
\centering
    \begin{tabular}{cccc}
    \hspace{-2mm} \includegraphics[width=0.11\textwidth]{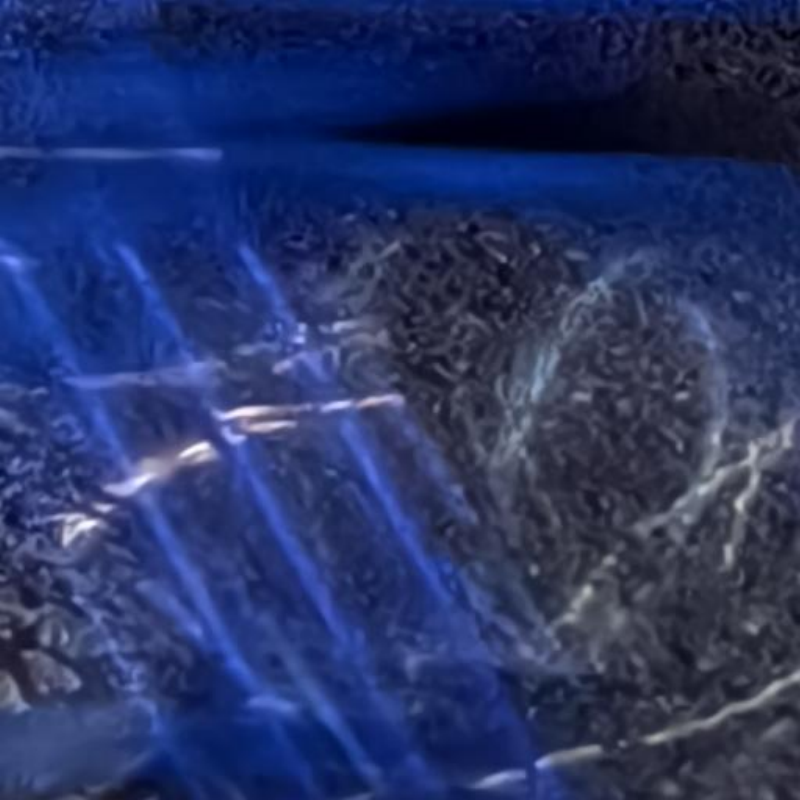}
    & \hspace{-3mm} \includegraphics[width=0.11\textwidth]{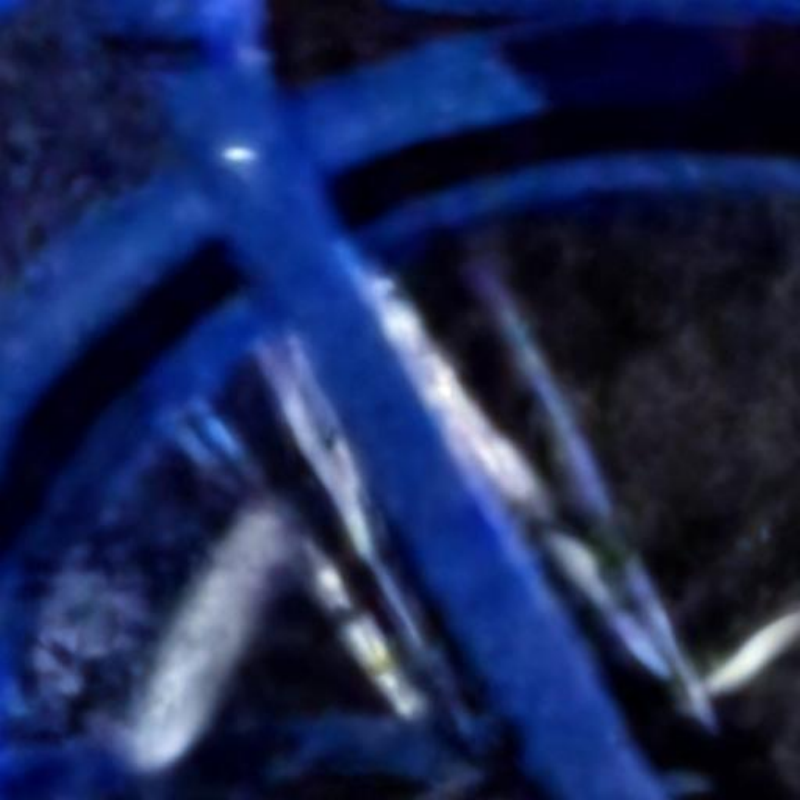}
    & \hspace{-3mm} \includegraphics[width=0.11\textwidth]{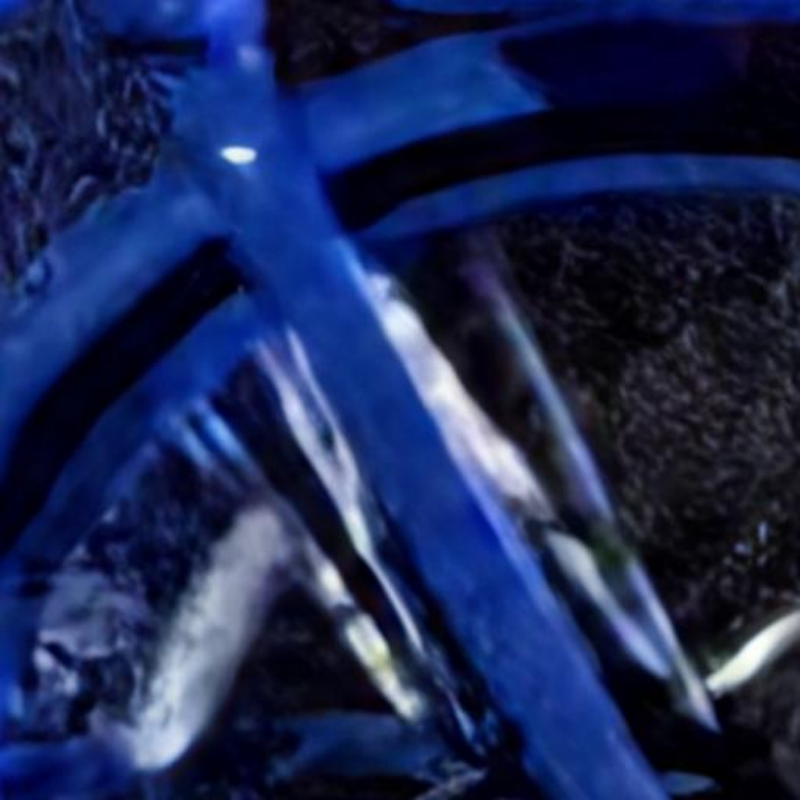}
    & \hspace{-3mm} \includegraphics[width=0.11\textwidth]{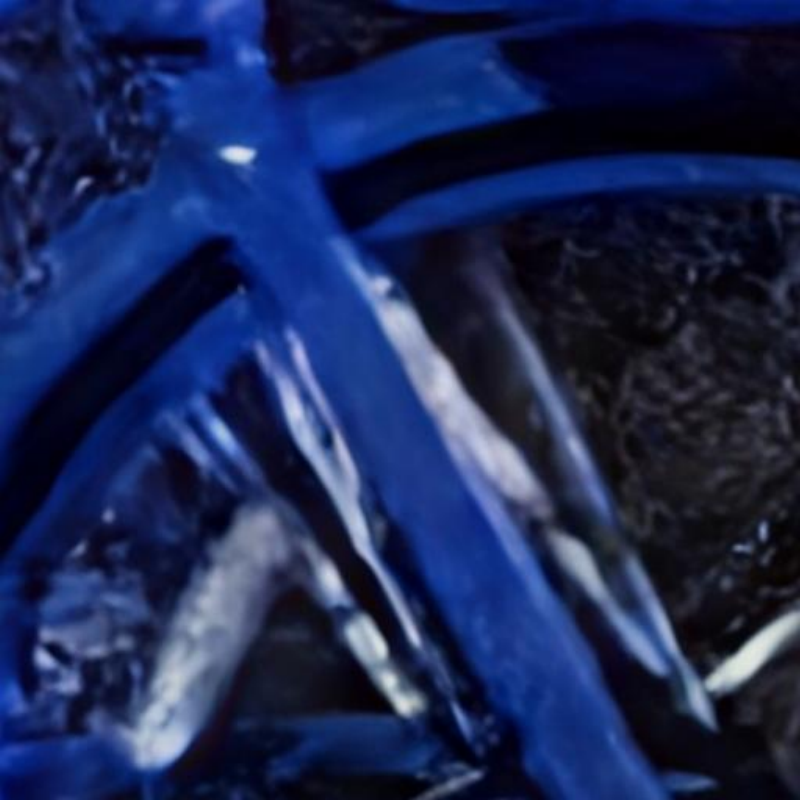} \\
    \hspace{-2mm}(a) Blurry input &\hspace{-3.7mm} (b) ${\text{w/~} M^{c}}$  &\hspace{-5.1mm} (c) ${\text{w/~} M}$ &\hspace{-5.2mm} (d) ${\text{w/~} M^{gt}}$\\
    \end{tabular}
\vspace{-2mm}
    \caption{Effectiveness of the proposed blurry-aware guided image restoration on image deblurring. (b) illustrates the restoration results using the method with $M^{c}$, which partially enhances image details but still leaves some residual blur.}
    \label{fig:ablation3}
\vspace{-3mm}
\end{figure}

{\flushleft \textbf{Effect of the context-based local blur detection module.}}
The proposed context-based local blur detection module to improve the identification of blurry regions.
To demonstrate the effectiveness of the proposed context-based blurred regions detection module in our proposed method, we first replace the $M$ with ground truth mask $M^{gt}$ (${\text{w/ $M$}^{gt}}$ for short).
The result in Table~\ref{tab:abation3} shows that using the mask generated by the context-based blurred regions detection module (${\text{w/~} M}$ for short) performs Competitive with the ground truth one.
Then, we demonstrate the effectiveness of $M$ on \ournet.
We replace the $M$ with not been binarized confidence map (${\text{w/~} M^{c}}$ for short).
Table~\ref{tab:abation3} shows that the proposed method achieves better results, where the MUSIQ-PAQ2PIQ value is 0.67 better.

\begin{figure}[t]
\footnotesize
\centering
    \begin{tabular}{cccc}
    \hspace{-2mm} \includegraphics[width=0.11\textwidth]{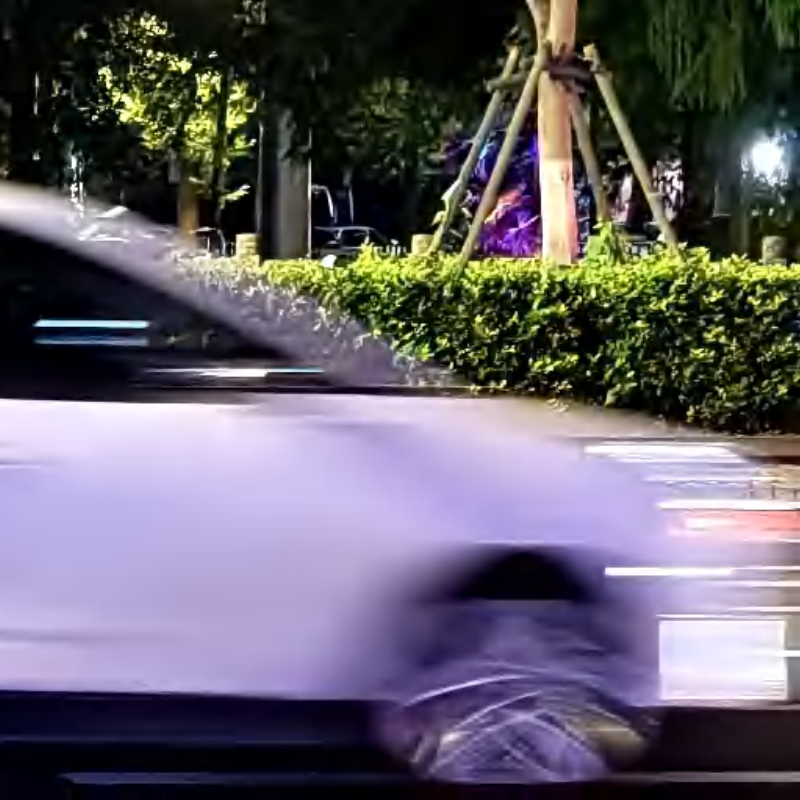}
    & \hspace{-3mm} \includegraphics[width=0.11\textwidth]{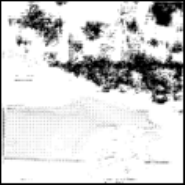}
    & \hspace{-3mm} \includegraphics[width=0.11\textwidth]{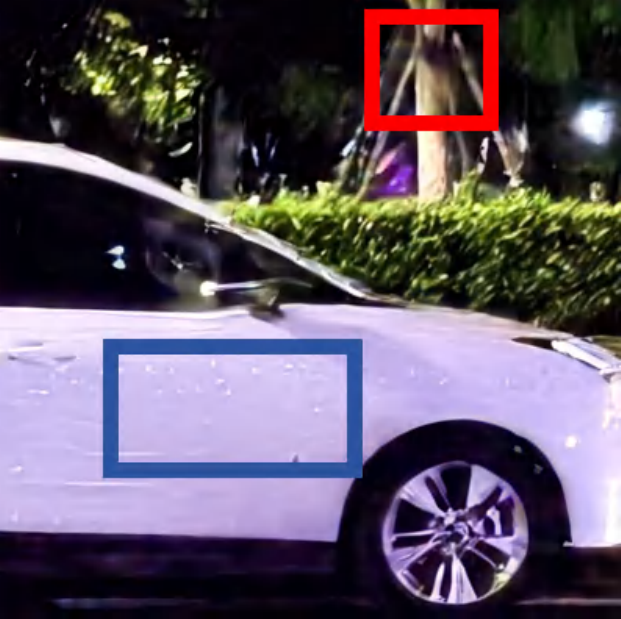}
    & \hspace{-3mm} \includegraphics[width=0.11\textwidth]{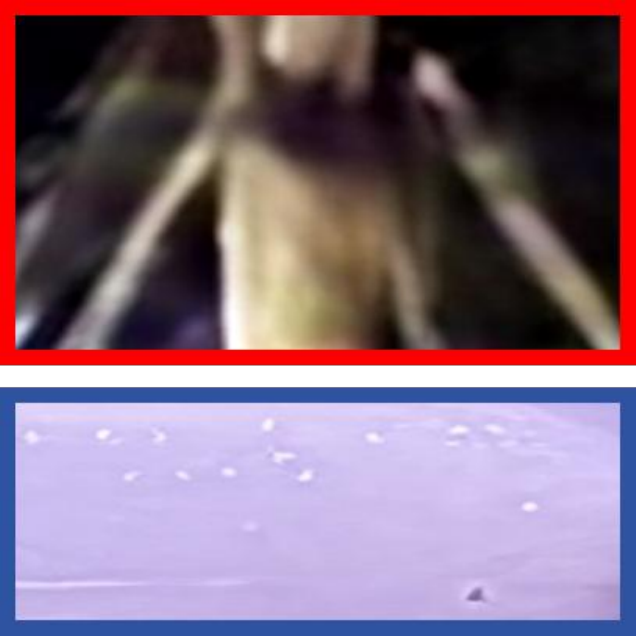} \\
    \hspace{-2mm}(a)  &\hspace{-3.7mm} (b)  &\hspace{-5.1mm} (c)  &\hspace{-5.2mm} (d) \\

    \hspace{-2mm} \includegraphics[width=0.11\textwidth]{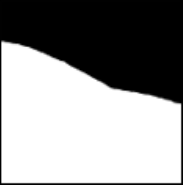}
    & \hspace{-3mm} \includegraphics[width=0.11\textwidth]{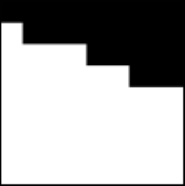}
    & \hspace{-3mm} \includegraphics[width=0.11\textwidth]{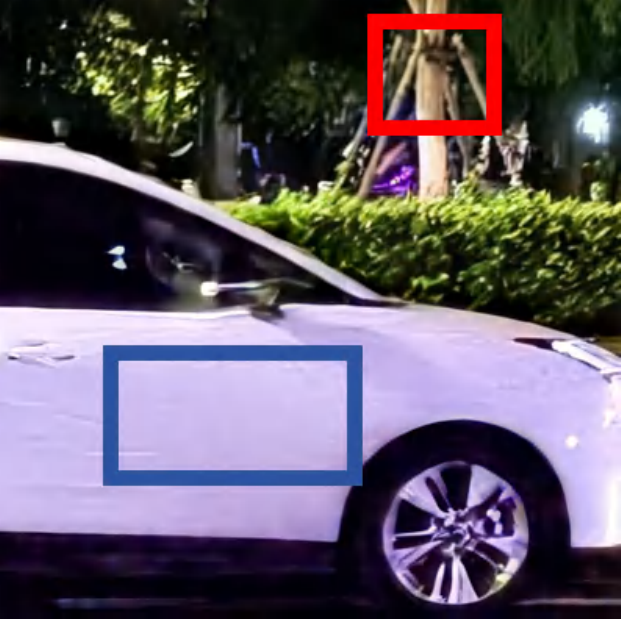}
    & \hspace{-3mm} \includegraphics[width=0.11\textwidth]{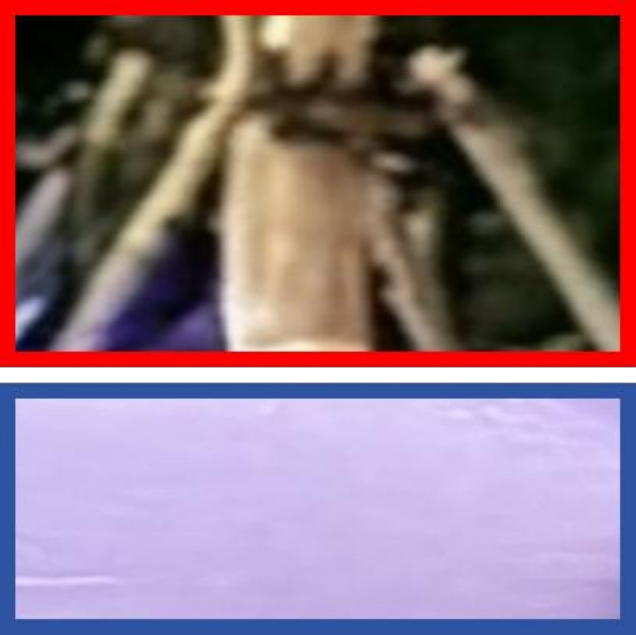} \\
    \hspace{-2mm}(e)  &\hspace{-3.7mm} (f)  &\hspace{-5.1mm} (g)  &\hspace{-5.2mm} (h) \\
    \end{tabular}
    \vspace{-1mm}

    \caption{Effectiveness of the context-based local blur detection module in blur region recognition. (a) Blurry input; (b) blurry region identification result of the context-based local blur detection module trained to identify for each pixel blur or not; (c) result of our method generated using the mask in (b); (d) zoomed-in details of the red and blue regions in (c); (e) ground truth mask; (f) blurry region identification result of the context-based local blur detection module; (g) result of our method generated using the mask in (f); (h) zoomed-in details of the red and blue regions in (g).}


    \label{fig:arxiv2}
    \vspace{-5mm}
\end{figure}

The visual comparisons in Figure~\ref{fig:ablation3}(c) demonstrate that using the context-based local blur detection module generates results that are competitive compared to the use of ground truth masks.
We further compare the context-based local blur detection module and the pixel-wise based local blur detection method in terms of their ability to identify blur regions.
The method based on pixel information does not identify the blur regions well as shown in Figure~\ref{fig:arxiv2}(b).
Figure~\ref{fig:arxiv2}(f) shows that the context-based local blur detection module better identifies blur regions and ultimately helps the model generate better results.

\section{Conclusion}
We propose ExpRDiff to explore features from short-exposure images and blurry regions to guide local blur removal and realistic image restoration.
To improve the identification of blurry regions, we design a context-based local blur detection module that incorporates additional contextual information to determine whether the regions are blurry or not.
Subsequently, we develop a blur-aware guided image restoration method that utilizes sharp structural details from short-exposure images, facilitating accurate reconstruction of heavily blurred regions.
Furthermore, to achieve realistic images with better visually appealing restoration, we present a short-exposure guided diffusion model that utilizes features from both short-exposure images and blurred regions to effectively guide the diffusion process.
Experimental results show that \ournet performs favorably against the state-of-art methods.

{
    \small
    \bibliographystyle{ieeenat_fullname}
    \bibliography{main}

\begin{thebibliography}{29}
\providecommand{\natexlab}[1]{#1}
\providecommand{\url}[1]{\texttt{#1}}
\expandafter\ifx\csname urlstyle\endcsname\relax
  \providecommand{\doi}[1]{doi: #1}\else
  \providecommand{\doi}{doi: \begingroup \urlstyle{rm}\Url}\fi

\bibitem[Blau and Michaeli(2018)]{tpdp}
Yochai Blau and Tomer Michaeli.
\newblock The perception-distortion tradeoff.
\newblock In \emph{CVPR}, 2018.

\bibitem[Chen et~al.(2008)Chen, Yuan, Tang, and Quan]{RDMD}
Jia Chen, Lu Yuan, Chi{-}Keung Tang, and Long Quan.
\newblock Robust dual motion deblurring.
\newblock In \emph{CVPR}, 2008.

\bibitem[Chen et~al.(2022)Chen, Chu, Zhang, and Sun]{NAFNet}
Liangyu Chen, Xiaojie Chu, Xiangyu Zhang, and Jian Sun.
\newblock Simple baselines for image restoration.
\newblock In \emph{ECCV}, 2022.

\bibitem[Cho et~al.(2021)Cho, Ji, Hong, Jung, and Ko]{MIMO}
Sung-Jin Cho, Seo-Won Ji, Jun-Pyo Hong, Seung-Won Jung, and Sung-Jea Ko.
\newblock Rethinking coarse-to-fine approach in single image deblurring.
\newblock In \emph{ICCV}, 2021.

\bibitem[He et~al.(2010)He, Sun, and Tang]{GIF}
Kaiming He, Jian Sun, and Xiaoou Tang.
\newblock Guided image filtering.
\newblock In \emph{ECCV}, 2010.

\bibitem[Jang et~al.(2017)Jang, Gu, and Poole]{gumbelsoftmax}
Eric Jang, Shixiang Gu, and Ben Poole.
\newblock Categorical reparameterization with gumbel-softmax.
\newblock In \emph{ICLR}, 2017.

\bibitem[Kong et~al.(2023)Kong, Dong, Ge, Li, and Pan]{Fftformer}
Lingshun Kong, Jiangxin Dong, Jianjun Ge, Mingqiang Li, and Jinshan Pan.
\newblock Efficient frequency domain-based transformers for high-quality image
  deblurring.
\newblock In \emph{CVPR}, 2023.

\bibitem[Kupyn et~al.(2018)Kupyn, Budzan, Mykhailych, Mishkin, and
  Matas]{DeblurGAN}
Orest Kupyn, Volodymyr Budzan, Mykola Mykhailych, Dmytro Mishkin, and Jiri
  Matas.
\newblock Deblurgan: Blind motion deblurring using conditional adversarial
  networks.
\newblock In \emph{CVPR}, 2018.

\bibitem[Kupyn et~al.(2019)Kupyn, Martyniuk, Wu, and Wang]{DeblurGANv2}
Orest Kupyn, Tetiana Martyniuk, Junru Wu, and Zhangyang Wang.
\newblock Deblurgan-v2: Deblurring (orders-of-magnitude) faster and better.
\newblock In \emph{ICCV}, 2019.

\bibitem[Li et~al.(2023)Li, Zhang, Jiang, Luo, Feng, and Xu]{ReLoBlur}
Haoying Li, Ziran Zhang, Tingting Jiang, Peng Luo, Huajun Feng, and Zhihai Xu.
\newblock Real-world deep local motion deblurring.
\newblock In \emph{AAAI}, 2023.

\bibitem[Li et~al.(2024)Li, Zhao, Zhou, Feng, Li, and Loy]{LMD-ViT}
Haoying Li, Jixin Zhao, Shangchen Zhou, Huajun Feng, Chongyi Li, and
  Chen~Change Loy.
\newblock Adaptive window pruning for efficient local motion deblurring.
\newblock In \emph{ICLR}, 2024.

\bibitem[Lin et~al.(2023)Lin, He, Chen, Lyu, Dai, Yu, Ouyang, Qiao, and
  Dong]{DiffBIR}
Xinqi Lin, Jingwen He, Ziyan Chen, Zhaoyang Lyu, Bo Dai, Fanghua Yu, Wanli
  Ouyang, Yu Qiao, and Chao Dong.
\newblock Diffbir: Towards blind image restoration with generative diffusion
  prior.
\newblock \emph{arXiv preprint arXiv:2308.15070}, 2023.

\bibitem[Loshchilov and Hutter(2019)]{adamw}
Ilya Loshchilov and Frank Hutter.
\newblock Decoupled weight decay regularization.
\newblock In \emph{ICLR}, 2019.

\bibitem[Mao et~al.(2021)Mao, Liu, Shen, Li, and Wang]{Deeprft}
Xintian Mao, Yiming Liu, Wei Shen, Qingli Li, and Yan Wang.
\newblock Deep residual fourier transformation for single image deblurring.
\newblock \emph{CoRR}, abs/2111.11745, 2021.

\bibitem[Mou et~al.(2024)Mou, Wang, Xie, Wu, Zhang, Qi, and Shan]{T2I_adapter}
Chong Mou, Xintao Wang, Liangbin Xie, Yanze Wu, Jian Zhang, Zhongang Qi, and
  Ying Shan.
\newblock T2i-adapter: Learning adapters to dig out more controllable ability
  for text-to-image diffusion models.
\newblock In \emph{AAAI}, 2024.

\bibitem[Podell et~al.(2024)Podell, English, Lacey, Blattmann, Dockhorn,
  M{\"{u}}ller, Penna, and Rombach]{SDXL}
Dustin Podell, Zion English, Kyle Lacey, Andreas Blattmann, Tim Dockhorn, Jonas
  M{\"{u}}ller, Joe Penna, and Robin Rombach.
\newblock {SDXL:} improving latent diffusion models for high-resolution image
  synthesis.
\newblock In \emph{ICLR}, 2024.

\bibitem[Suin et~al.(2020)Suin, Purohit, and Rajagopalan]{SAPHN}
Maitreya Suin, Kuldeep Purohit, and A.~N. Rajagopalan.
\newblock Spatially-attentive patch-hierarchical network for adaptive motion
  deblurring.
\newblock In \emph{CVPR}, 2020.

\bibitem[Tao et~al.(2018)Tao, Gao, Shen, Wang, and Jia]{SRN}
Xin Tao, Hongyun Gao, Xiaoyong Shen, Jue Wang, and Jiaya Jia.
\newblock Scale-recurrent network for deep image deblurring.
\newblock In \emph{CVPR}, 2018.

\bibitem[Tsai et~al.(2022)Tsai, Peng, Lin, Tsai, and Lin]{Stripformer}
Fu-Jen Tsai, Yan-Tsung Peng, Yen-Yu Lin, Chung-Chi Tsai, and Chia-Wen Lin.
\newblock Stripformer: Strip transformer for fast image deblurring.
\newblock In \emph{ECCV}, 2022.

\bibitem[Vaswani et~al.(2017)Vaswani, Shazeer, Parmar, Uszkoreit, Jones, Gomez,
  Kaiser, and Polosukhin]{Transformer}
Ashish Vaswani, Noam Shazeer, Niki Parmar, Jakob Uszkoreit, Llion Jones,
  Aidan~N. Gomez, Lukasz Kaiser, and Illia Polosukhin.
\newblock Attention is all you need.
\newblock In \emph{NIPS}, 2017.

\bibitem[Wang et~al.(2022)Wang, Cun, Bao, Zhou, Liu, and Li]{Uformer}
Zhendong Wang, Xiaodong Cun, Jianmin Bao, Wengang Zhou, Jianzhuang Liu, and
  Houqiang Li.
\newblock Uformer: A general u-shaped transformer for image restoration.
\newblock In \emph{CVPR}, 2022.

\bibitem[Yang et~al.(2023)Yang, Wu, Ren, Xie, and Zhang]{PASD}
Tao Yang, Rongyuan Wu, Peiran Ren, Xuansong Xie, and Lei Zhang.
\newblock Pixel-aware stable diffusion for realistic image super-resolution and
  personalized stylization.
\newblock \emph{arXiv preprint arXiv:2308.14469}, 2023.

\bibitem[Yu et~al.(2024)Yu, Gu, Li, Hu, Kong, Wang, He, Qiao, and Dong]{SUPIR}
Fanghua Yu, Jinjin Gu, Zheyuan Li, Jinfan Hu, Xiangtao Kong, Xintao Wang,
  Jingwen He, Yu Qiao, and Chao Dong.
\newblock Scaling up to excellence: Practicing model scaling for
  photo-realistic image restoration in the wild.
\newblock In \emph{CVPR}, 2024.

\bibitem[Yuan et~al.(2007)Yuan, Sun, Quan, and Shum]{IDBNIP}
Lu Yuan, Jian Sun, Long Quan, and Heung{-}Yeung Shum.
\newblock Image deblurring with blurred/noisy image pairs.
\newblock \emph{{ACM} Trans. Graph.}, 26, 2007.

\bibitem[Zamir et~al.(2021)Zamir, Arora, Khan, Hayat, Khan, Yang, and
  Shao]{MPRNet}
Syed~Waqas Zamir, Aditya Arora, Salman Khan, Munawar Hayat, Fahad~Shahbaz Khan,
  Ming-Hsuan Yang, and Ling Shao.
\newblock Multi-stage progressive image restoration.
\newblock In \emph{CVPR}, 2021.

\bibitem[Zamir et~al.(2022)Zamir, Arora, Khan, Hayat, Khan, and
  Yang]{Restormer}
Syed~Waqas Zamir, Aditya Arora, Salman Khan, Munawar Hayat, Fahad~Shahbaz Khan,
  and Ming-Hsuan Yang.
\newblock Restormer: Efficient transformer for high-resolution image
  restoration.
\newblock In \emph{CVPR}, 2022.

\bibitem[Zhang et~al.(2019)Zhang, Dai, Li, and Koniusz]{DMPHN}
Hongguang Zhang, Yuchao Dai, Hongdong Li, and Piotr Koniusz.
\newblock Deep stacked hierarchical multi-patch network for image deblurring.
\newblock In \emph{CVPR}, 2019.

\bibitem[Zhang et~al.(2023)Zhang, Rao, and Agrawala]{controlnet}
Lvmin Zhang, Anyi Rao, and Maneesh Agrawala.
\newblock Adding conditional control to text-to-image diffusion models.
\newblock In \emph{ICCV}, 2023.

\bibitem[Zhuo et~al.(2010)Zhuo, Guo, and Sim]{RFD}
Shaojie Zhuo, Dong Guo, and Terence Sim.
\newblock Robust flash deblurring.
\newblock In \emph{CVPR}, 2010.

\end{thebibliography}
}


\end{document}